%% file: preprint.tex
\documentclass{article}

\newif\ifshowgenerations
\showgenerationstrue

 \usepackage{iclr2026_conference,times}

\usepackage[utf8]{inputenc}
\usepackage[T1]{fontenc}
\usepackage{hyperref}
\hypersetup{
  pdfborder={0 0 2},
  linkbordercolor={0.55 0.80 0.79},
  citebordercolor={0.55 0.80 0.79},
  urlbordercolor={0.55 0.80 0.79},
}
\usepackage{url}
\usepackage{booktabs}
\usepackage{amsfonts}
\usepackage{amssymb}
\usepackage{nicefrac}
\usepackage{microtype}
\usepackage{xcolor}
\usepackage{graphicx}
\usepackage{float}
\usepackage{tcolorbox}
\usepackage{amsmath}
\usepackage{wrapfig2}
\usepackage{xfrac}
\usepackage{bm}
\usepackage{xspace}
\usepackage[labelfont=bf]{caption}

\iftrue
\newcommand{\cutsectionup}{\vspace*{-8pt}}
\newcommand{\cutsectiondown}{\vspace*{-4pt}}
\newcommand{\cutsubsectionup}{\vspace*{-8pt}}
\newcommand{\cutsubsectiondown}{\vspace*{-4pt}}

\newcommand{\cutparagraphup}{\vspace*{-7pt}}

\newcommand{\cutcaptionup}{\vspace*{-8pt}}
\newcommand{\cutcaptiondown}{\vspace*{-8pt}}
\newcommand{\cutwrapfigup}{\vspace*{-10pt}}

\newcommand{\cutequationdown}{\vspace*{-0pt}}

\newcommand{\cutfloatsep}{\setlength{\textfloatsep}{14pt plus 2pt minus 4pt}%
  \setlength{\intextsep}{10pt plus 2pt minus 2pt}%
  \setlength{\floatsep}{10pt plus 2pt minus 2pt}}
\newcommand{\cutfindingpad}{1mm}
\newcommand{\cutdisplayskip}{\setlength{\abovedisplayskip}{7pt plus 2pt minus 4pt}%
  \setlength{\belowdisplayskip}{7pt plus 2pt minus 4pt}%
  \setlength{\abovedisplayshortskip}{0pt plus 2pt}%
  \setlength{\belowdisplayshortskip}{4pt plus 2pt minus 2pt}}
\else
\newcommand{\cutsectionup}{\vspace*{-0pt}}
\newcommand{\cutsectiondown}{\vspace*{-0pt}}
\newcommand{\cutsubsectionup}{\vspace*{-0pt}}
\newcommand{\cutsubsectiondown}{\vspace*{-0pt}}

\newcommand{\cutparagraphup}{\vspace*{-0pt}}

\newcommand{\cutcaptionup}{\vspace*{-0pt}}
\newcommand{\cutcaptiondown}{\vspace*{-0pt}}
\newcommand{\cutwrapfigup}{\vspace*{-0pt}}

\newcommand{\cutequationdown}{\vspace*{-0pt}}

\newcommand{\cutfloatsep}{\setlength{\textfloatsep}{20pt plus 2pt minus 4pt}%
  \setlength{\intextsep}{12pt plus 2pt minus 2pt}%
  \setlength{\floatsep}{12pt plus 2pt minus 2pt}}
\newcommand{\cutfindingpad}{2mm}
\newcommand{\cutdisplayskip}{\setlength{\abovedisplayskip}{10pt plus 2pt minus 5pt}%
  \setlength{\belowdisplayskip}{10pt plus 2pt minus 5pt}%
  \setlength{\abovedisplayshortskip}{0pt plus 3pt}%
  \setlength{\belowdisplayshortskip}{6pt plus 3pt minus 3pt}}
\fi

\cutfloatsep
\AtBeginDocument{\cutdisplayskip}

\newcommand{\abra}{\textsc{Abra}\xspace}
\newcommand{\abrasize}[1]{\textsc{Abra}-#1}

\newcommand{\omittedfig}{\fbox{\parbox[c][3cm][c]{0.85\linewidth}{\centering\itshape [generation figure omitted; enable \textbackslash showgenerationstrue for the full version]}}}

\newcommand{\figscale}{0.82}

\definecolor{findingbody}{RGB}{250, 236, 247}
\tcbset{colback=findingbody, boxrule=0.75pt, arc=2pt, width=\columnwidth, center,
        top=\cutfindingpad, bottom=\cutfindingpad}

\newenvironment{finding}{\begin{tcolorbox}\textbf{Finding:} }{\end{tcolorbox}}

\makeatletter
\g@addto@macro\UrlBreaks{\do\-\do\_\do\.\do\/\do\:\do\=\do\?\do\&\do\1\do\2\do\3\do\4\do\5\do\6\do\7\do\8\do\9\do\0}
\makeatother

\title{\abra: Scaling Diffusion Image Training}

\author{%
\textbf{Kyle Chickering, Wei-An Lin, Swayam Bhanded, Dan Saunders,}\\
\textbf{Akshat Tripathi, Jiaming Song, Shyamal Buch, Xinchen Yan}\\
Luma AI}

\iclrfinalcopy

\begin{document}

\maketitle
\lhead{Preprint.}

\begin{abstract}
    Compute-optimal scaling laws guide the training of frontier language models yet remain largely unexplored for visual generation. We present a systematic scaling law study for text-to-image diffusion models using \abra, a controlled family of flow-matching transformers trained across three orders of magnitude worth of compute ($10^{19}$ to $10^{22}$ FLOPs), reaching significantly larger compute budgets than previous works. We demonstrate that diffusion models scale just as predictably as language models but require far more data to train optimally: compute optimality occurs at approximately $200$ image tokens per parameter, ten times the Chinchilla compute-optimal prescription for LLMs. We show that unlike language models, diffusion models are robust to overtraining and that practitioners should err on the side of more data rather than a larger model. Finally, we show that this predictability extends beyond training loss to generative quality metrics, optimal CFG settings, representation quality, and even the shape of the training curves, which collapse onto a universal form.
\end{abstract}

\input{content/01_introduction}
\input{content/02_related_work}
\input{content/03_experimental_setup}
\input{content/04_empirical_results}
\input{content/05_conclusion}

\subsubsection*{Acknowledgments}
We thank Terrance DeVries and Linqi (Alex) Zhou for helpful discussion throughout the completion of this work.

\newpage
\bibliographystyle{iclr2026_conference}
\bibliography{references}

\newpage
\appendix

\input{content/a01_architecture}
\input{content/a02_limitations}
\input{content/a03_heterogeneous}
\input{content/a04_model_generations}
\input{content/a05_extended_scaling_analysis}
\input{content/a06_fit_procedures}
\input{content/a07_parameter_counting}

\end{document}

%% file: content/01_introduction.tex
\begin{figure}[H]
    \centering
    \includegraphics[width=\figscale\textwidth]{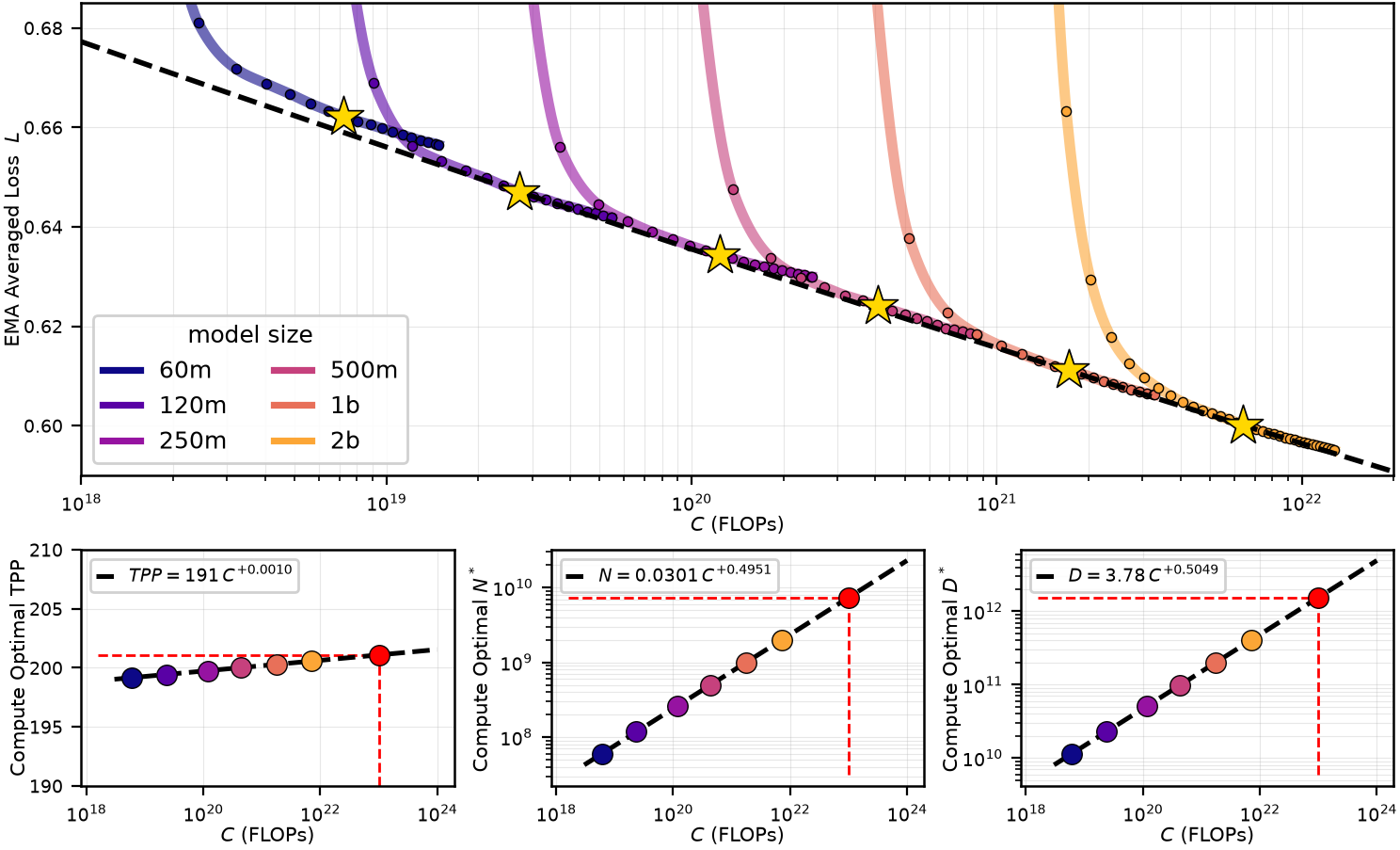}
    \cutcaptionup
    \caption{Text-to-image diffusion transformers follow predictable scaling laws and require ten times as much data as a comparably sized language model to reach compute optimality.}
    \cutcaptiondown
    \label{fig:headline}
\end{figure}

\cutsectionup
\section{Introduction}
\cutsectiondown

Compute-optimal scaling laws describe how a fixed training compute budget $C$ (in FLOPs) should be allocated between model size $N$ (in parameters) and dataset size $D$ (in training tokens) in order to minimize loss~\citep{hestness2017deep,amodei2016deep,kaplan2020scaling,hoffmann2022empirical}. For large language models compute should be divided evenly between the model size and the dataset size according to $N\propto C^{1/2}$ and $D\propto C^{1/2}$, giving the ``Chinchilla rule'' of training LLMs to $20$ tokens per parameter (TPP)~\citep{hoffmann2022empirical}. Scaling analysis has become a standard industrial tool in language modeling and is used to guide the training of frontier models~\citep{touvron2023llama,openai2023gpt4,deepseek2024llm,grattafiori2024llama3,sun2024hunyuan,kimiteam2025kimik2,microsoftai2026mai}.

The extension of scaling law analysis to the visual generation domain presents challenges not present in the language domain. Visual content has higher dimensionality (i.e., 2D images), resolution dependent token information density, separate training and generation paradigms, and much noisier training curves. The scaling of these models remains understudied despite the release of many large-scale visual generation studies. Existing work by~\citet{liang2024scaling} fits iso-FLOP profiles below $10^{19}$ FLOPs and extrapolates to $10^{21}$ FLOPs. We use an order of magnitude more compute and a more controlled family which allows us to derive much more precise empirical fits.

We significantly refine the scaling foundations for text-to-image models by conducting a systematic scaling study using \abra, a controlled family of dense flow-matching transformers ranging from 60M to 2B parameters. We utilize $\mu$P to fit scaling laws spanning three orders of magnitude (from $10^{19}$ to $10^{22}$ FLOPs). Our central findings indicate that compute optimality is achieved at roughly $200$ image tokens per parameter for text-to-image diffusion modeling, a $10\times$ increase over the standard Chinchilla rule for LLMs. Furthermore, we find that diffusion model training is robust to overtraining, giving practitioners confidence to err on the side of training a smaller model with more data.

We fit compute-optimal scaling laws to the evaluation loss and a range of common generative metrics, finding that all of these quantities scale predictably. We study the scaling of model representation quality using linear probing and find that representation quality also follows predictable scaling trends. We ablate the effect of resolution on our scaling laws and we demonstrate that like LLMs, diffusion models exhibit scaling collapse.

Our main contributions are:
\begin{itemize}
    \item We present a comprehensive and controlled study of compute-optimal scaling for dense text-to-image transformers and derive an \textbf{actionable 200 TPP} rule for training diffusion models.
    \item We show that diffusion model training is \textbf{robust to overtraining} and precisely characterize this phenomenon.
    \item We extend our analysis to show that common generative metrics scale predictably, and that \textbf{representation and generation capabilities scale heterogeneously}.
    \item We show for the first time that diffusion models exhibit \textbf{scaling collapse}.
\end{itemize}

The remainder of the paper is organized as follows. We review related work on neural scaling laws and diffusion models in Section~\ref{sec:related}, describe our experimental setup and scaling ladder in Section~\ref{sec:setup}, and present our empirical findings in Section~\ref{sec:results}.

%% file: content/02_related_work.tex
\cutsectionup
\section{Related Work}\label{sec:related}
\cutsectiondown

\paragraph*{Text-to-Image Transformers and Flow-Matching.}
\citet{peebles2023scalable} introduced and studied the efficacy of vision transformers in the context of generative diffusion modeling~\citep{sohldickstein2015deep}. The original diffusion formulations~\citep{ho2020ddpm,song2021scorebased,song2021ddim} have largely been replaced by flow-matching~\citep{lipman2023flow,liu2023rectified,albergo2023building,heitz2023iterative} which formulates the denoising process as simple linear interpolation. We use $v$-prediction introduced by \citet{albergo2023building} and \citet{ma2024sit} and the latent diffusion formulation~\citet{rombach2022latent} and \citet{kingma2014vae}.

\cutparagraphup
\paragraph*{Neural Scaling Laws.}

While the origins of neural scaling laws trace their roots to the nineties (see~\citealp{caballero2023broken}), their modern invocation and study can be largely attributed to the works~\citep{amodei2016deep,hestness2017deep,kaplan2020scaling,hoffmann2022empirical}. These works find that many machine learning models, and large language models in particular, exhibit predictable improvements in performance as a function of compute and model size. While there is an extensive body of literature studying the scaling behaviors of language models, for example~\citep{bergsma2025power,bergsma2025scaling,tao2024scaling,alabdulmohsin2022revisiting,muennighoff2023scaling,dey2023cerebras}, comparatively little is known about the scaling of diffusion models.

\cutparagraphup
\paragraph*{Scaling Laws for Diffusion Models.}
As previously mentioned,~\citet{liang2024scaling} perform what is, to our knowledge, the only scaling law study for text-to-image diffusion modeling. Our work uses an order of magnitude more compute and a controlled model family to get refined estimates of the compute-optimal scaling laws for text-to-image diffusion models. \citet{liang2024scaling} find that model size should be scaled faster than dataset size, which leads to systematic undertraining at frontier scales. We find that data and parameters should be scaled at the same rate, i.e., $D\propto C^{1/2}$ and $N\propto C^{1/2}$.

A related line of work~\citep{li2024scalability, li2024efficient} studies the scalability of text-to-image models on downstream benchmarks and metrics, but does not perform a compute-optimal analysis. \citet{mei2025bigger} study how latent diffusion models behave as capacity and compute grow but focus on inference scaling and do not do compute-optimal analysis. Their study systematically explores how optimal CFG changes with model size and we corroborate their finding regarding CFG and model size in the compute-optimal setting. \citet{yan2025rethinking} show that pixel-sequence generative models reach compute optimality at roughly $200$--$400$ TPP. For video models,~\citet{yin2025towards} study the scaling properties of video training but only fit scaling laws using models up to 250M parameters. Finally, the works~\citep{zheng2025scaling, ryu2026summer22b} apply $\mu$P~\citep{yang2021tuning} to diffusion transformers to enable proxy tuning and hyperparameter transfer.

\citet{ge2026chimera} is a concurrent work which also studies compute-optimal scaling in the context of image and video generation. Despite large architectural differences between our model family and theirs the two works agree in the diagnosis that diffusion model training requires more data than LLM training. However the two works differ substantially in the types of questions which they answer. We focus here on foundational questions about scaling image generation and our work provides extensive evaluation of generative metrics, a study of compute-optimal representation quality, analysis of how image resolution affects the scaling of diffusion models, and the first validation of scaling collapse phenomena for diffusion models. These aspects of scaling image generation are not within the scope of \citet{ge2026chimera}.

%% file: content/03_experimental_setup.tex
\cutsectionup
\section{Method}\label{sec:setup}
\cutsectiondown
We aim to characterize how text-to-image diffusion transformers should be scaled: for a given training budget, how many parameters should we use and how much data should we train on? This section describes our approach to answering these questions.

\cutsubsectionup
\subsection{Compute Optimality}\label{sec:compute_optimality}
\cutsubsectiondown
We follow the definition of compute optimality from~\citet{hoffmann2022empirical}: given a fixed training budget of $C$ FLOPs, we seek the model size $N$ and dataset size $D$ minimizing a loss or evaluation metric $\mathcal{L}$,
\begin{align}\label{eq:compute_optimal}
    N_{\text{opt}}(C),\, D_{\text{opt}}(C) = \arg\min_{N, D} \,\mathcal{L}(N, D; C).
\end{align}
\cutequationdown
Throughout this paper, $N$ denotes the total parameter count of the diffusion transformer, excluding the frozen text encoder.
Unlike in LLMs, the embedding and unembedding layers contribute a negligible fraction of parameters at our
scales, so we report total parameters throughout: including or excluding them shifts the optimal-TPP estimate by less
than $3\%$.
$D$ denotes the number of image tokens seen during training. We measure data in \emph{image}
tokens per parameter, abbreviated \textup{TPP}, and write \textup{MTPP} for \emph{multimodal}
tokens per parameter when text and image tokens are counted together.
To locate this compute-optimal point accurately, we intentionally overtrain models in the \abra family so that the optimum $(N_{\text{opt}}, D_{\text{opt}})$ lands on the interior of our search range.
We train up to $400$ TPP, an upper bound informed by our prior experience training diffusion models and by the autoregressive pixel sequence results of~\citet{yan2025rethinking}.

\begin{figure}
    \centering
    \ifshowgenerations
        \includegraphics[width=0.7\textwidth]{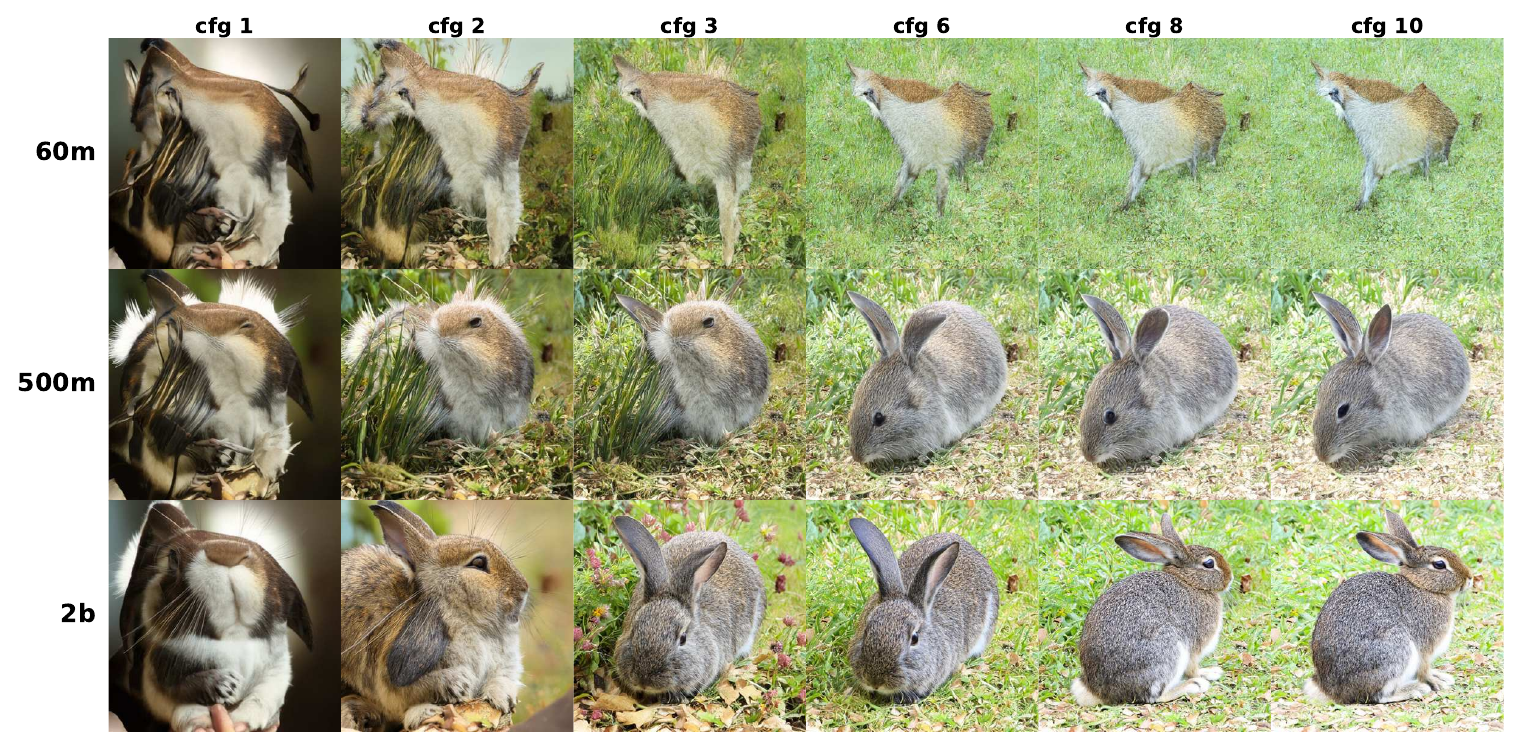}
    \else
        \omittedfig
    \fi
    \cutcaptionup
    \caption{$512\times 512$ generations of the prompt \texttt{``a rabbit''} across model scales and CFG.
    }
    \cutcaptiondown
    \label{fig:cfg_marquee}
\end{figure}

\cutsubsectionup
\subsection{Training and Sampling}\label{sec:flow_objective}
\cutsubsectiondown
We train latent diffusion models with a $v$-prediction flow-matching objective~\citep{ma2024sit, salimans2022progressive}.
Specifically, the network $v_\theta(x_t, t; \texttt{text})$ regresses the velocity of the linear interpolant between a clean image latent and Gaussian noise.
We maintain an EMA of the model weights with decay parameter $0.9999$~\citep{ho2020ddpm, peebles2023scalable}.
Because we have training steps roughly fixed we keep the EMA timescale constant across model sizes~\citep{karras2024edm2}.
We train on a base resolution of $512\times 512$ with five image aspect-ratio buckets.
Each image is encoded by the frozen $f=8$ FLUX VAE into a $64 \times 64$
latent, which a parameter-free $2 \times 2$ fold reduces to a $32 \times 32$
grid that the transformer consumes directly.
To make the unconditional estimate available at inference, we drop text conditioning with probability 0.1 during training.

We integrate the ODE $\frac{\mathrm{d} x_t}{\mathrm{d} t} = v_{\theta}(x_t, t; \texttt{text})$ backwards in time starting from $\varepsilon \sim \mathcal{N}(0, 1)$ at $t=1$ with a 50-step Euler solver, sampling in the latent space and decoding to pixels with the FLUX autoencoder~\citep{blackforestlabs2025flux2}.
We apply classifier-free guidance~\citep{ho2022classifier} with guidance scale swept per model and held fixed across sampling steps.

\cutsubsectionup
\subsection{Model Family}\label{sec:model_family}
\cutsubsectiondown
We construct \abra, a family of diffusion transformers from $60$M to $2$B parameters, with depth growing from $19$ to $33$ blocks and width from $384$ to $1792$, spanning roughly $10^{19}$ to $10^{22}$ image-token FLOPs. Our models combine aspects from various open source text-to-image models into a new family which is amenable to scaling~\citep{blackforestlabs2025flux2,chen2024pixart}.

Each model uses four double-stream blocks followed by single-stream blocks for the remainder of the layers, with an FFN expansion rate of $6$. \citet{liang2024scaling} found that a constant aspect ratio yields smooth iso-FLOP curves. For \abra we use a monotonically non-decreasing width-to-depth ratio under the assumption that this smoothness persists under monotone aspect ratios.
Text conditioning is provided by a frozen Qwen3-4B model~\citep{yang2025qwen3} and we exclude the conditioner parameters from our FLOP and model size counts.
Across the ladder we double the batch size every time we double the parameter count, keeping the number of training steps fixed.

A full accounting of the \abra architecture can be found in Appendix~\ref{app:arch}.

\cutsubsectionup
\subsection{Choice of Dataset}
\cutsubsectiondown
We train on the DataComp-1B dataset~\citep{gadre2023datacomp}, a large-scale collection of web image-text pairs. Because web-sourced captions are often low quality, we re-caption the images using a variety of open-source vision-language models, following~\citep{betker2023dalle3}. During training we include a mix of dense, medium, sparse, and web-sourced captions. We do not perform mid or post training of our models and focus exclusively on diffusion model pre-training. We report metrics on either DataComp-1B or on a held-out evaluation mixture that is more diverse than the training data, of which DataComp-1B is a $\sim\!20\%$ subset.

\cutsubsectionup
\subsection{$\mu$P and Hyperparameter Selection}\label{sec:mup}
\cutsubsectiondown

We transfer hyperparameters across the model family with $\mu$P~\citep{yang2021tuning}, using the Complete-P parameterization~\citep{dey2025dont}, and train with the Adam optimizer~\citep{kingma2015adam}.
We validate our implementation using spectral coordinate checking~\citep{chickering2026gqa}. We validate $\mu$-transfer for \abrasize{60M}, \abrasize{120M}, and \abrasize{250M}, and then zero-shot transfer the optimal learning rate, $4\times 10^{-4}$, to the remaining models in the ladder.
We sweep the initialization scale on the \abrasize{60M} model and initialize all layers from the same value of $\sigma=0.01$.
Unlike~\citet{mlodozeniec2025completed} and~\citet{jiang2026hyperparameter}, we do not tune per-layer base hyperparameters.
We use a logit-normal timestep sampler with mean $p_\mu=1.9$ and standard deviation $p_\sigma=1.0$.

%% file: content/04_empirical_results.tex
\cutsectionup
\section{Empirical Results}\label{sec:results}
We now turn to the scaling analysis itself. Our central finding is that text-to-image flow-matching transformers scale predictably across a wide variety of evaluative metrics, but require roughly ten times as much data per parameter as LLMs to reach compute optimality. $N$ denotes the parameter count of the vision transformer and $D$ is measured in image tokens.

\begin{figure}[!h]
    \centering
    \includegraphics[width=\figscale\textwidth]{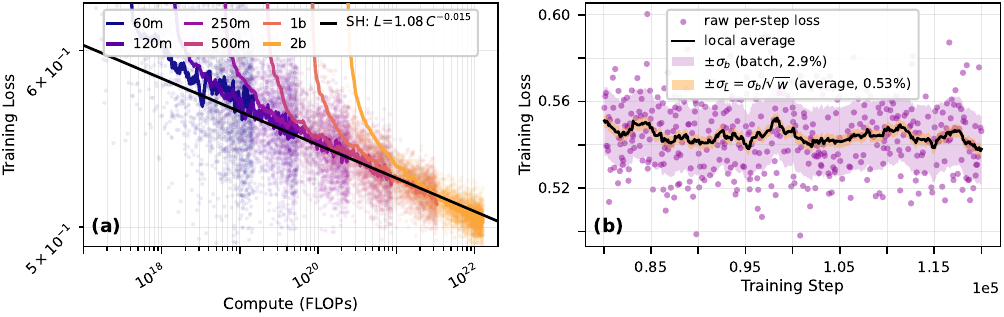}
    \cutcaptionup
    \caption{\emph{Left:} The raw training loss for \abra runs. The bold curves are a simple moving average of the scattered loss data. The solid black line is a qualitative fit of this data. \emph{Right:} The noise decomposition for the \abrasize{250M} run, illustrating that the per-step differences in the loss curve are dominated by the batch-to-batch noise.}
    \cutcaptiondown
    \label{fig:raw_loss}
\end{figure}

\cutsubsectionup
\subsection{Estimating Compute-Optimal Tokens per Parameter}\label{subsec:tpp_estimation}
\cutsubsectiondown

\begin{tcolorbox}
\textbf{Finding:} Diffusion model training is compute-optimal at $200$ TPP, roughly $10\times$ higher than the canonical LLM compute-optimal budget of $20$ TPP.
\end{tcolorbox}

Figure~\ref{fig:raw_loss}(a) shows the training loss for the \abra{} family with a qualitative fit of the data. As we can see from Figure~\ref{fig:raw_loss}(b) the raw per-step loss is dominated by batch noise. Over 10{,}000 steps, the batch-to-batch noise on the \abrasize{250M} run is roughly $4$ to $7\times$ larger than the improvement in the reducible loss over the same period.
All fits in this section will therefore target the EMA model loss (\textbf{not} an EMA of the training loss as was done in~\citep{liang2024scaling}) which is substantially smoother than the raw model loss.

\begin{wrapfigure}{r}{0.48\textwidth}
    \cutwrapfigup
    \centering
    \includegraphics[width=0.46\textwidth]{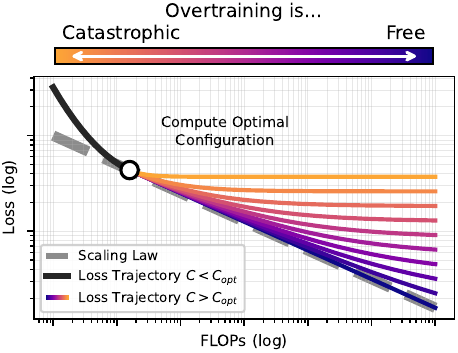}
    \caption{Overtraining compute-optimal models incurs a cost at fixed compute. The compute-optimal point occurs when the model loss curve is tangent to the scaling law curve.}
    \label{fig:scaling_divergence}
\end{wrapfigure}

To construct these curves without training hundreds of models we exploit the constant LR schedule together with the smoothness of the EMA loss and interpolate each model's evaluation checkpoints into a continuous loss-versus-TPP trajectory using PCHIP interpolation
\citep{fritsch1980monotone,fritsch1984method}. The evaluation loss is computed on our evaluation dataset by taking 100{,}000 samples and averaging the loss over the logit-normal timestep distribution using the EMA weight predictions.
At each TPP $T$ we fit a power law in compute, $L(C; T) = A_T C^{-\alpha_T} + \beta_T$; fixed-$C$ slices of this surface give the iso-FLOP profiles, and the $\arg\min$ over $T$ gives the compute-optimal TPP.
The minima of these curves tightly cluster around 200 TPP at all but the smallest of our training budgets. The relative flatness of these profiles is also conspicuous and we investigate this point further in Section~\ref{subsec:overtraining}.
We exclude \abrasize{60M} as a clear outlier (Figure~\ref{fig:headline}), in line with the observation that scaling laws hold only in an appropriate regime~\citep{hestness2017deep}.

There are a multitude of ways to fit scaling laws (see~\citealp{li2025misfitting} for a survey) and to ensure our findings are reliable we conduct an in-depth comparison against many other fit procedures in Appendix~\ref{app:fit_procedures} with our findings tabulated in Table~\ref{tab:procedure_comparison}. Across every fit procedure we tried, the optimal TPP lies within a narrow band of only about $\pm 17$ TPP.

\begin{figure}[!b]
    \centering
    \includegraphics[width=\figscale\textwidth]{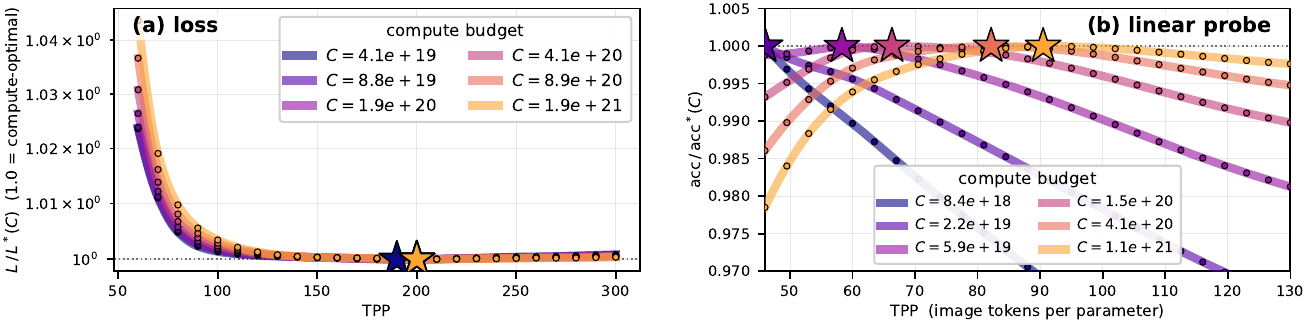}
    \cutcaptionup
    \caption{\textbf{Normalized iso-FLOP curves.} \emph{(a)} EMA loss $L/L^*(C)$ and \emph{(b)} linear-probe accuracy $\mathrm{acc}/\mathrm{acc}^*(C)$ versus TPP for several compute budgets, each normalized to its per-budget optimum (star). The near-flat curves show the compute-optimal TPP is largely budget-insensitive; representational quality (b) is optimal at a substantially lower TPP than the loss (a).}
    \cutcaptiondown
    \label{fig:isoflop_combined}
\end{figure}

\cutsubsectionup
\subsection{Diffusion Overtraining is Forgiving}\label{subsec:overtraining}
\cutsubsectiondown

\begin{tcolorbox}
    \textbf{Finding:} Overtraining a diffusion model is nearly free (in the sense of Figure~\ref{fig:scaling_divergence}).
\end{tcolorbox}

In practice, models are frequently trained well past their compute-optimal point. The most common reason is inference cost: a smaller model trained on more data is cheaper to serve, so practitioners deliberately overtrain to shift compute from deployment to training~\citep{touvron2023llama, gadre2025language}.

Therefore, scaling law analysis cannot simply end at the compute-optimal point. It is important to understand precisely how we expect the model to behave when overtrained. Since the compute-optimal point is definitionally a minimum over the loss~\eqref{eq:compute_optimal}, any deviation from compute optimality necessarily incurs some cost (in training FLOPs). This idea is illustrated visually in Figure~\ref{fig:scaling_divergence}. Concretely, fix a compute budget $C$ and let $N_{\text{opt}}(C)$ denote the compute-optimal parameter count at that budget. For a model with $N \ne N_{\text{opt}}$ parameters trained at the same compute $C$ (equivalently, at a non-optimal TPP), we can define the \textbf{loss penalty}
\begin{align}
    \Delta L(N, C) := L(N, C) - L\big(N_{\text{opt}}(C),\, C\big). \label{eq:penalty}
\end{align}
The quantity $\Delta L$ should be thought of as the additional loss we incur by \emph{not} training the compute-optimal model size for a budget of $C$ FLOPs.

\begin{figure}[!ht]
    \centering
    \includegraphics[width=\figscale\textwidth]{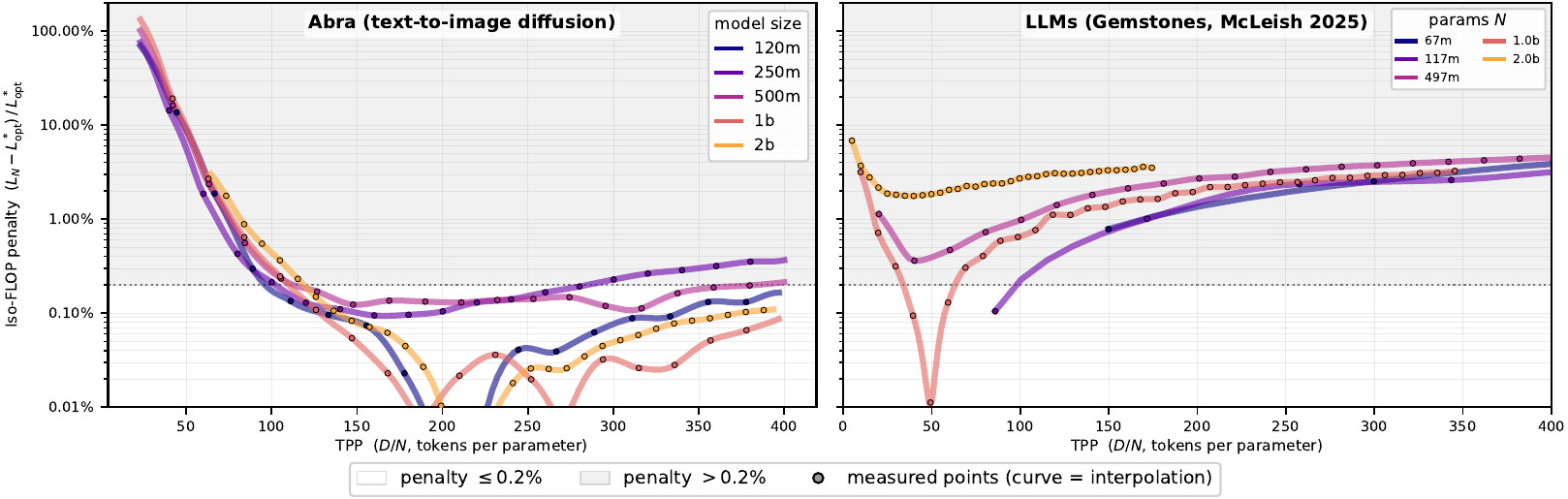}
    \cutcaptionup
    \caption{Relative iso-FLOP loss penalty $\Delta L / L^*$ (Eq.~\eqref{eq:penalty}) as a function of TPP for \emph{(left)} the \abra text-to-image diffusion family and \emph{(right)} the Gemstones LLM suite~\citep{mcleish2025gemstones}. \abra markers denote evaluated checkpoints and Gemstones curves are the raw per-checkpoint fineweb-edu validation-loss penalties. The shaded bands mark the $0.2\%$ penalty threshold (white below, grey above).}
    \cutcaptiondown
    \label{fig:avg_loss_penalty}
\end{figure}

Figure~\ref{fig:avg_loss_penalty} plots this loss penalty as a relative percentage, $\Delta L(N, C) / L\big(N_{\text{opt}}(C),\, C\big)$, as a function of TPP for the \abra family. We additionally plot the same data for an open-source LLM scaling family, Gemstones~\citep{mcleish2025gemstones}, which spans the parameter range $48$M--$2$B and is heavily (and purposefully) overtrained. We provide additional comparisons against other overtrained LLM studies in Appendix~\ref{app:extended_scaling} (Figures~\ref{fig:gadre_penalty},~\ref{fig:penalty_chinchilla},~and~\ref{fig:penalty_porian}).

From Figure~\ref{fig:avg_loss_penalty} it is clear that diffusion models suffer almost no loss penalty when $2\times$ overtrained (less than $0.5\%$). In the parlance of Figure~\ref{fig:scaling_divergence}, overtraining diffusion models is nearly ``free''. This is manifestly not the case for LLMs, where the loss penalty is substantial even for modest overtraining (see Figure~\ref{fig:avg_loss_penalty}, right).

On the other hand, Figure~\ref{fig:avg_loss_penalty} also reveals that \textit{undertraining} diffusion models is catastrophic. This observation yields a practical rule: in compute-limited settings it is far safer to err on the side of overtraining a small model than risk undertraining a large one.

\cutsubsectionup
\subsection{Generative Metrics Scale Predictably}\label{sec:generative_metrics}
\cutsubsectiondown

\begin{finding}
Generative metrics follow power laws in compute but their compute-optimal allocations diverge.
\end{finding}

Low diffusion loss does not necessarily indicate strong generative performance~\citep{theis2016note,ho2020ddpm}. Generation quality for diffusion models is therefore measured using distributional level metrics computed on the outputs of the ODE sampling procedure. Common metrics include FID~\citep{heusel2017gans}, KID~\citep{binkowski2018demystifying}, and CLIPScore~\citep{hessel2021clipscore}, and all of these metrics are sensitive to the classifier-free guidance (CFG) scale. Different guidance strengths favor different qualitative aspects of image generation. High guidance typically adheres better to prompts and low guidance typically adheres better to the target data distribution~\citep{ho2022classifier,saharia2022imagen}. Because generative quality depends so heavily on the choice of CFG it is insufficient to compare models at a fixed guidance scale. Instead, we must find the optimal guidance scale for every model on each metric.

\begin{figure}[tbp]
    \centering
    \includegraphics[width=\figscale\textwidth]{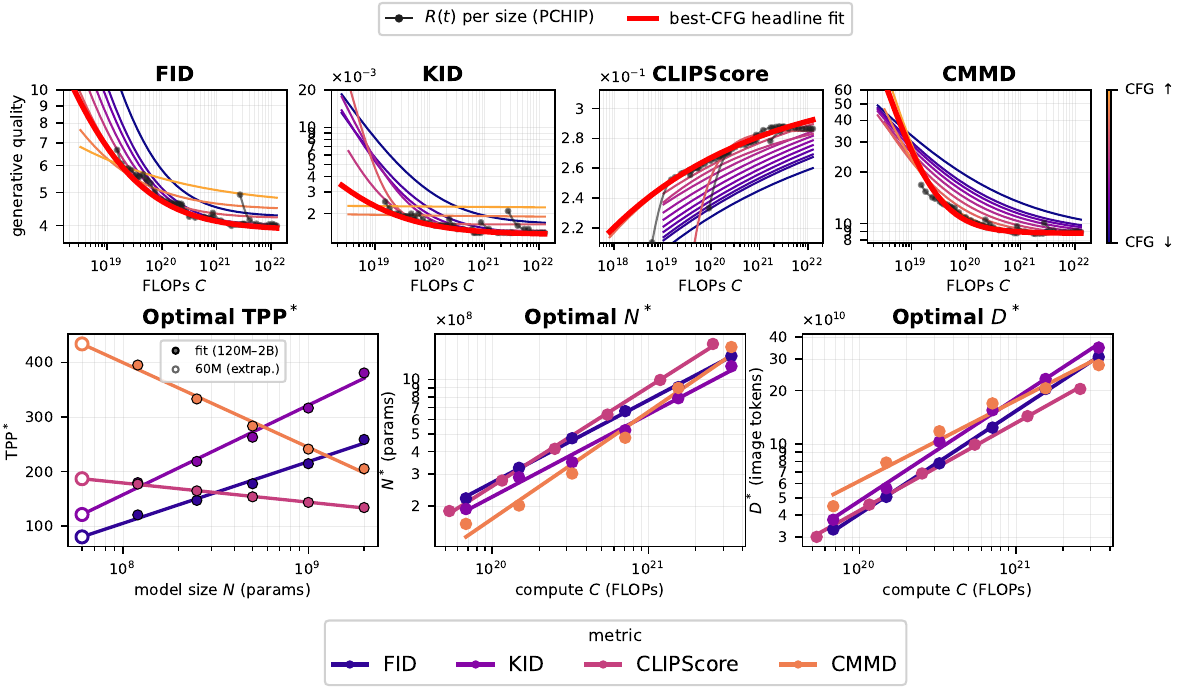}
    \cutcaptionup
    \caption{\textbf{Standard generative metrics scale predictably.} \emph{Top:} generative quality vs training compute $C$ for FID, KID, CLIPScore, and CMMD. The bold red curve is its compute-optimal frontier. \emph{Bottom:} compute-optimal $\mathrm{TPP}^*$ as a function of model size $N$, and the compute-optimal $N_{\text{opt}}$ and $D_{\text{opt}}$ as functions of compute $C$. }
    \cutcaptiondown
    \label{fig:standard_metrics}
\end{figure}

To this end, we sweep the optimal guidance scale for each model size and every metric. We generate $50{,}000$ samples for FID and $25{,}000$ samples for all other metrics. Figure~\ref{fig:standard_metrics}~(top) shows that every tested metric follows a power law of the form from Section~\ref{subsec:tpp_estimation}. We fit $M(C) = A\,C^{-\alpha} + F_M$, with the sign of the power term flipped for
CLIPScore.
The offset $F_M$ plays the role of the loss law's $\beta_T$ and Table~\ref{tab:metric_fits} collects the fitted coefficients.

Figure~\ref{fig:standard_metrics}~(bottom) shows that compute-optimal allocation between $D$ and $N$ differs sharply across metrics. FID and KID require the dataset size to grow much more rapidly than the model size to reach compute optimality. On the other hand, CLIPScore and CMMD show decreasing optimal TPP as a function of compute, indicating that the number of parameters should grow more rapidly than the dataset size. We must therefore conclude that no single metric can define compute optimality, mirroring the situation where no single metric can determine visual quality~\citep{stein2023exposing,jayasumana2024rethinking}. We note that the heterogeneous capacity demands of these common generative metrics warrants further study into how these metrics correlate with human preferences.

Finally, we find that the optimal CFG decreases with both model size and training steps.
\citet{mei2025bigger} report the same trend for model size and for sampling steps, but did not investigate this trend over training horizons.
In summary, stronger generative capacity needs less guidance and the feature model behind each metric also shifts the optimal CFG. Appendix~\ref{app:metric_fits} gives our
full CFG analysis (Figures~\ref{fig:cfg_sweeps} and~\ref{fig:optimal_cfg_flops}), including analyzing a DINOv2~\citep{oquab2024dinov2} family of metrics: FDD, KDD, and DMMD.

\cutsubsectionup
\subsection{Linear Probe Accuracy}
\cutsubsectiondown
\label{subsec:linear_probes}
\begin{figure}[tbp]
    \centering
    \includegraphics[width=\figscale\textwidth]{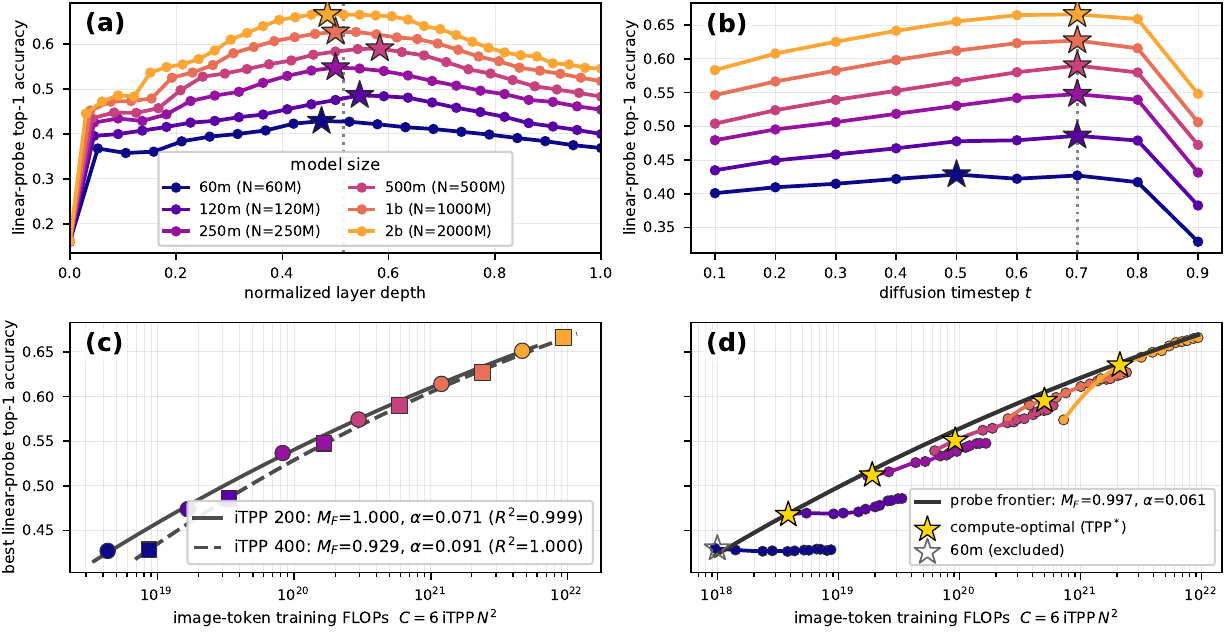}
    \cutcaptionup
    \caption{\textbf{(a)} Best top-1 accuracy as a function of relative layer depth. \textbf{(b)} Best top-1 accuracy as a function of diffusion timestep. \textbf{(c)} Top-1 accuracy follows predictable scaling laws at different TPP. \textbf{(d)} Top-1 accuracy as a function of training compute.}
    \cutcaptiondown
    \label{fig:probe_layer}
\end{figure}

\begin{tcolorbox}
    \textbf{Finding:} Linear-probe accuracy follows predictable scaling laws. Understanding performance is compute-optimal at far lower TPP than generation.
\end{tcolorbox}

Diffusion models are not only good visual generators, they are also good representation learners~\citep{xiang2023denoising}. To investigate how the model's understanding performance scales as a function of compute we conduct linear probing experiments~\citep{chen2020igpt,alain2016understanding,xiang2023denoising} on frozen activations from each \abra{} checkpoint and report ImageNet~\citep{deng2009imagenet} top-1 accuracy. Following~\citet{xiang2023denoising}, we pool per-block features from a forward pass at a fixed diffusion timestep, and we sweep layers, timesteps, and training steps.

Figures~\ref{fig:probe_layer}(a) and~(b) show that linear probing accuracy peaks for middle layers (consistent with~\citep{chen2020igpt,yan2025rethinking,xiang2023denoising}) and for the moderately high diffusion timestep $t\approx 0.7$. Figure~\ref{fig:probe_layer}(c) demonstrates that linear-probe accuracy $a^*(C) = \max_{t,\ell} \mathrm{acc}(C; t, \ell)$ follows tight scaling laws at fixed TPP.

Additionally, we seek to understand the compute-optimal configuration for image understanding. Figure~\ref{fig:isoflop_combined}(b) applies the same iso-FLOP analysis of Section~\ref{subsec:tpp_estimation} to the per-model best probe accuracies $a^*(C)$ shown in Figure~\ref{fig:probe_layer}(d). The optimal TPP for image understanding is much lower than the TPP for generation in the compute range that we tested, but notably the optimal allocation is far from the balanced $D\propto N$ that we found for generation quality. For image understanding optimal TPP is steadily increasing, indicating that FLOPs are more efficiently used for increasing dataset size rather than model size. The finding that understanding is easier than generation at these scales is consistent with the works of~\citet{yan2025rethinking} and~\citet{xiang2023denoising}.

This observation suggests that there is a ``double-point'' where both generation and understanding are jointly compute-optimal. We estimate this point to occur when training with about $5\times 10^{22}$ FLOPs and a 6B parameter model. Models larger than this, trained to compute optimality for generation, will no longer be undertrained for understanding.

\cutsubsectionup
\subsection{Scaling Collapse}\label{subsec:collapse}
\cutsubsectiondown

\begin{tcolorbox}
    \textbf{Finding:} Compute optimal diffusion model training exhibits scaling collapse.
\end{tcolorbox}

\begin{figure}[tbp]
    \centering
    \includegraphics[width=\figscale\textwidth]{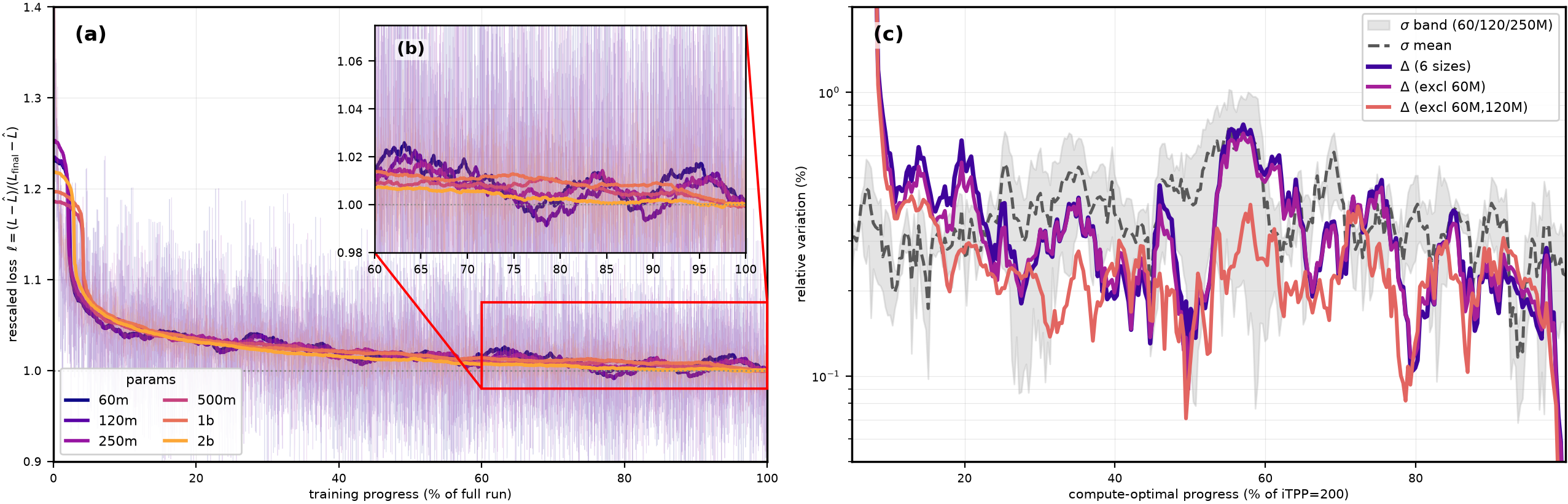}
    \cutcaptionup
    \caption{\textbf{(a)} Training loss curves for six models from the \abra family rescaled with Equation~\eqref{eq:collapse_rescale}. \textbf{(b)} The deviation $\Delta$ between models collapses beneath the noise floor $\sigma$ for the three seed-replicated models (\abrasize{60M}, \abrasize{120M}, and \abrasize{250M}).}
    \cutcaptiondown
    \label{fig:collapse}
\end{figure}

Conventional scaling law analysis fits power laws in quantitative metrics \emph{after training}~\citep{hestness2017deep, kaplan2020scaling, hoffmann2022empirical}. While~\citet{kaplan2020scaling} attempt to fit the loss curves themselves to a sum of power laws, these fits do not seem to satisfactorily capture the qualitative aggregate scaling behavior of the training dynamics. More recently~\citet{qiu2025scaling} showed that the loss curves of compute-optimally trained language models follow a universal, non-power-law form. Training loss curves, when appropriately rescaled, exhibit \textbf{scaling collapse} where the training curve for each individual model in the scaling ladder collapse onto a single trajectory. Practically speaking the differences (in rescaled loss) between models of different sizes are \emph{smaller} than the noise from running a single model with different seeds. This enables practitioners to use scaling collapse as a diagnostic tool during training and ensure that models remain near compute-optimal. To our knowledge this phenomenon has not been examined for diffusion models until now.

Following~\citet{qiu2025scaling} we let $p$ index our model family and consider the rescaled compute $C(p) = xt^*(p)$, where $x \in [0, 1]$ and $t^*(p)$ is the compute used for model $p$. Thus $x$ represents the percentage of training that has finished. We then define the rescaled loss
\begin{align}\label{eq:collapse_rescale}
    \ell(x, p; \theta) = \frac{L(x\,t^*(p),\, p) - \hat L}{L(t^*(p),\, p) - \hat L},
\end{align}
as well as the \emph{collapse deviation} $\Delta$ and the per-model (relative) noise floor
\begin{align*}
    \Delta(x) := \frac{\mathbb{V}_{p, \theta(p)}[\ell(x, p; \theta)]^{1/2}}{\mathbb{E}_{p, \theta(p)}[\ell(x, p; \theta)]}, \qquad
    \sigma(x; p) := \frac{\mathbb{V}_{\theta(p)}[\mathcal{L}(xt^*(p), p, \theta(p))]^{1/2}}{\mathbb{E}_{\theta(p)}[\mathcal{L}(xt^*(p), p, \theta(p))]},
\end{align*}
where $\mathcal{L}$ is the shifted loss $\mathcal{L}=L-\hat{L}$ and where $\hat{L}$ is the irreducible loss for the model family, found via our fit procedure. Rescaling in this manner fixes the endpoints of the training curves to match so that collapse indicates that the shape of the training curves depends only on relative compute. We run the \abrasize{60M}, \abrasize{120M}, and \abrasize{250M} with four different seeds, which change both the initialization and the data ordering. We cut off the data window at a fixed $200$ TPP for all of the models based on the analysis above.

Figure~\ref{fig:collapse} shows the result of our experiment. We plot the raw-loss curves with an overlaid SMA (left) and the raw $\Delta, \sigma$ values with their own SMA (right); the moving average is required in the context of diffusion since the training loss curves, and thus $\Delta$ and $\sigma$, are extremely noisy. Because we only ran multiple seeds up to 250M parameters, we can only determine the collapse for these three models. Despite the noise, we see a clear collapse in training loss to the noise floor within the first 15\% of training. For the remainder of training, the cross-size deviation $\Delta$ remains at or below the noise floor $\sigma$, indicating that the \abra family satisfies the criteria for scaling collapse and is therefore compute-optimal (or nearly compute-optimal).

Our results are consistent with the findings of~\citet{qiu2025scaling} for models trained with a constant learning rate schedule. Of particular interest is our finding that collapse is a phenomenon which extends beyond the context of simple MLPs and LLMs. Despite being trained with a different loss target and data paradigm, compute-optimal collapse behavior persists.

\cutsubsectionup
\subsection{The Effect of Resolution}\label{subsec:resolution}
\begin{tcolorbox}
    \textbf{Finding:} Higher resolution training requires more image tokens to reach compute optimality.
\end{tcolorbox}
The slope of a scaling law is understood to be a property dependent more on the data distribution than the model architecture~\citep{hestness2017deep,bahri2024explaining,kaplan2020scaling}. The data distribution for images changes quite dramatically when we change resolution: the information per patch goes down and we rescale the timestep sampling distribution according to~\citet{hoogeboom2023simple}. Therefore the natural hypothesis is that as image resolution increases we would expect that the diffusion models will require an increasing number of image tokens to reach compute optimality.

We show that this hypothesis is correct. We train the \abrasize{120M} through \abrasize{500M} ladder at $256$, $384$, $512$, and $768$~pixel resolutions. For each model we use the~\citet{hoogeboom2023simple} SNR rescaling on our base distribution. Models are evaluated using the log-normally distributed timesteps and thus we cannot compare losses directly across resolutions. However, for all resolutions we continue to see scaling behavior as a function of compute.

Because our largest models are only 500M, getting clean scaling law fits is challenging. To this end, and motivated by our analysis above, we assume the balanced compute optimality constraint $a=b=0.5$. Under this assumption we find that the compute-optimal TPP systematically increases as a function of resolution from about $165$ at $256\times 256$ resolution to around $247$ at $768\times 768$ resolution. While the required image tokens per parameter increases, the number of images required to realize the larger data consumption actually goes down: we need fewer \textit{images} per-parameter as we increase resolution. This is consistent with the findings of~\citet{yan2025rethinking} in the context of pixel-autoregressive generation models.

\begin{figure}[tbp]
    \centering
    \includegraphics[width=\figscale\textwidth]{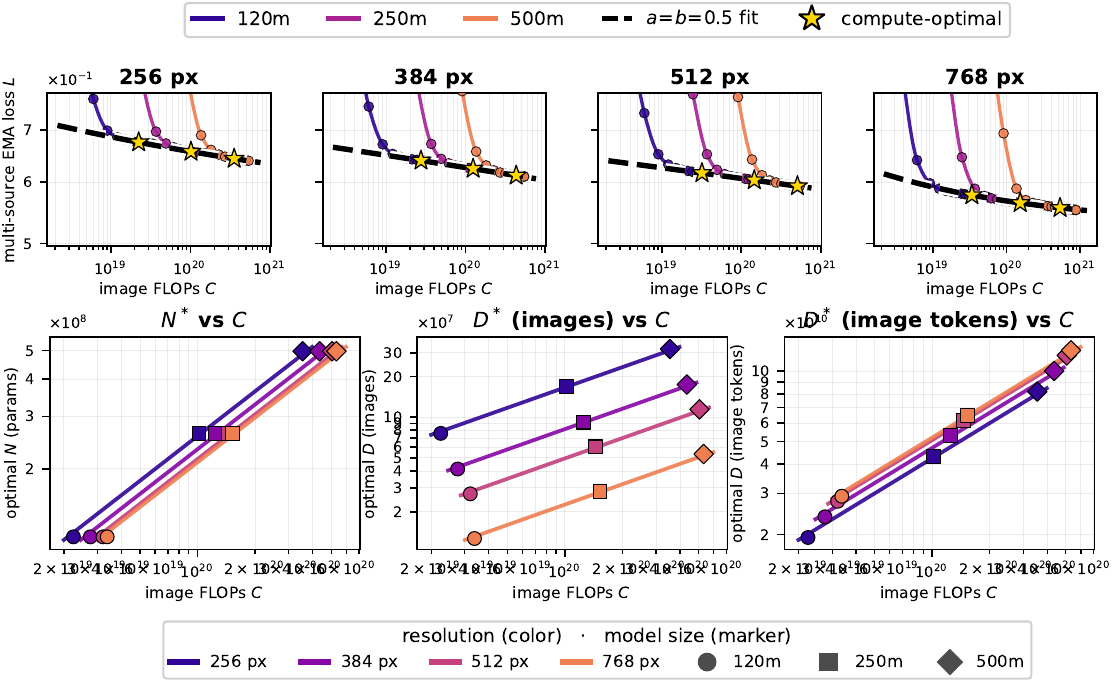}
    \cutcaptionup
    \caption{\textbf{(Top Row)} Training curves and loss scaling laws at different resolutions. \textbf{(Bottom Row)} TPP as a function of resolution, compute-optimal model size, compute-optimal \emph{number of images}, compute-optimal dataset size in number of image tokens. For these data we obtained estimated compute-optimal TPP of $165$, $202$, $235$, and $247$ respectively for the image sizes $256$, $384$, $512$, and $768$. }
    \cutcaptiondown
    \label{fig:resolution_placeholder}
\end{figure}

The results indicate that training end-to-end on high-resolution images is proportionally more costly in terms of FLOPs. One consequence is that high-resolution diffusion models can be trained compute-optimally on relatively few images, and high-resolution training is compute bound, not data bound.

%% file: content/05_conclusion.tex
\cutsectionup
\section{Conclusion}
\cutsectiondown
We revisited compute-optimal scaling for text-to-image diffusion pre-training with $10\times$ as much compute as previous studies and a controlled model family, \abra, which spans 60M to 2B parameters. Text-to-image diffusion transformers follow tight, predictable scaling laws with compute optimality achieved at approximately $200$ TPP, an order of magnitude higher than for LLMs.

We showed that diffusion model training is robust to overtraining, that common generative metrics like FID and CLIPScore follow predictable scaling trends in compute, and that as resolution increases the models take more tokens to reach compute optimality. Furthermore, we showed that the internal representation quality scales differently than generative capabilities and we showed, for the first time, that compute-optimal diffusion model training exhibits scaling collapse.

%% file: content/a01_architecture.tex
\cutsectionup
\section{Architectural Decisions}\label{app:arch}
\cutsectiondown
\textbf{Scaling Family.} Our architecture draws its components from various open source image generation models like FLUX.2~\citep{blackforestlabs2025flux2} and PixArt-$\alpha$~\citep{chen2024pixart}. We use ungated SiLU activation functions with an FFN expansion ratio of $6$ instead of the usual $4$. We fix the head dimension to $128$ and grow width through the number of heads. We use parametric RMSNorm~\citep{zhang2019root} as the QK-normalization strategy~\citep{henry2020query}. We use non-parametric RMSNorm for the layer normalization which we found to slightly outperform parametric RMSNorm in initial testing. We do a single AdaLN layer shared between all blocks~\citep{chen2024pixart}. We chose to use four double-stream blocks per model and scale depth by adding single-stream blocks. The per-model depth, width, head count, feed-forward expansion ratio, and training configuration are given in Table~\ref{tab:model_family}.

\begin{table}[!ht]
\centering
\cutcaptionup
\caption{The \abra model family and its training configuration. ``Double'' and ``Single'' count the double- and single-stream MMDiT blocks~\citep{esser2024sd3, blackforestlabs2025flux2}; ``Batch'' is the global batch size in images. Parameter counts exclude the frozen text encoder.}
\cutcaptiondown
\label{tab:model_family}
\small
\setlength{\tabcolsep}{4.5pt}
\begin{tabular}{lrrrrrrrr}
\toprule
Model & Params & Double & Single & Width & Heads & Batch (img.) & Steps & Image-tokens \\
\midrule
\abrasize{60M}  & $60.1$M  & $4$ & $15$ & $384$  & $3$  & $128$  & $183$k & $2.4\times10^{10}$ \\
\abrasize{120M} & $118.0$M & $4$ & $18$ & $512$  & $4$  & $256$  & $180$k & $4.7\times10^{10}$ \\
\abrasize{250M} & $262.3$M & $4$ & $18$ & $768$  & $6$  & $512$  & $200$k & $1.0\times10^{11}$ \\
\abrasize{500M} & $497.0$M & $4$ & $20$ & $1024$ & $8$  & $1024$ & $190$k & $2.0\times10^{11}$ \\
\abrasize{1B}   & $998.8$M & $4$ & $22$ & $1408$ & $11$ & $2048$ & $191$k & $4.0\times10^{11}$ \\
\abrasize{2B}   & $1.97$B  & $4$ & $29$ & $1792$ & $14$ & $4032$ & $191$k & $7.9\times10^{11}$ \\
\bottomrule
\end{tabular}
\end{table}

\textbf{Removal of Bias Terms and Weight Decay.}
We train our model with all bias terms removed. We found that this slightly improved performance in initial architectural ablations. This observation is not new and many other large-scale training efforts remove bias terms~\citep{chowdhery2023palm, touvron2023llama, karras2024edm2, blackforestlabs2025flux2, nvidia2025nemotronh}. We train our models with no weight decay~\citep{peebles2023scalable} which is standard in diffusion model training.

\textbf{Maximal Update Parameterization ($\mu$P).}
We apply $\mu$P to ensure transfer of optimal learning rate throughout the scaling ladder. We validated that transfer occurred for \abrasize{120M} and \abrasize{250M} of our scaling law experiments. We validated our $\mu$P implementation using spectral coordinate checking~\citep{chickering2026gqa,chickering2025mup}, which is more robust to catching implementation errors than the original formulation of coordinate checking from~\citet{yang2021tuning}.

\textbf{Softmax Scaling Factor.}
We found that using QK-norm together with $\mu$P occasionally led to instabilities in the training. We attributed these instabilities to the normalization scheme in the softmax. Using QK-norm changes the expected scaling of the softmax argument. In normal $\mu$P we assume that $Q, K\sim \mathcal{N}(0, 1/n)$, $x\sim \mathcal{N}(0, 1)$ from which we find that $Qx, Kx\sim \mathcal{N}(0, 1)$, with norms scaling like $\lVert Qx\rVert, \lVert Kx\rVert\sim \sqrt{n}$. However, with QK-norm, we instead have that $\lVert\widetilde{Q}x\rVert, \lVert\widetilde{K}x\rVert\sim 1$. Therefore, the products $(Qx)^{\top}(Kx)$ are expected to scale with $\sqrt{d}$ instead of just $d$. We validated this empirically by comparing the different softmax scalings in the presence of QK-norm.

\textbf{Epoch Repetition Factors.}
We report the number of epochs that each model is trained for in Table~\ref{tab:repetition}. At 400 TPP our 2B model trains for nearly 8 epochs. We note that diffusion models are known to be much more tolerant of repetition than LLMs~\citep{prabhudesai2025diffusion,muennighoff2023scaling}.
\begin{table}[!ht]
\centering
    \cutcaptionup
\caption{Epoch repetition factor (number of passes over the $100$M-sample mix) for
each model, $D_I/10^8$, where $D_I = B\cdot\text{steps}$ is the total images processed at
the final ($400$ TPP) checkpoint.}
    \cutcaptiondown
\label{tab:repetition}
\begin{tabular}{lccc}
\toprule
Model & Images $D_I$ & Epochs at 200 TPP & Epochs at 400 TPP \\
\midrule
\abrasize{60M}  & $23.4$M  & 0.12 & $0.23$ \\
\abrasize{120M} & $46.1$M  & 0.23 & $0.46$ \\
\abrasize{250M} & $102.4$M & 0.51 & $1.02$ \\
\abrasize{500M} & $194.6$M & 0.98 & $1.95$ \\
\abrasize{1B}   & $391.2$M & 1.95 & $3.91$ \\
\abrasize{2B}   & $770.1$M & 3.85 & $7.70$ \\
\bottomrule
\end{tabular}
\end{table}

%% file: content/a02_limitations.tex
\cutsectionup
\section{Limitations}\label{app:limitations}
\cutsectiondown
Our results rely on several axes remaining fixed. We speculate that ablations along these axes will result in a shift of the optimal TPP finding, as we saw already with resolution.

\begin{enumerate}
    \item \textbf{Optimizers:} We looked only at the Adam optimizer. However more recent optimizers like Muon~\citep{jordan2024muon}, Shampoo~\citep{gupta2018shampoo}, and SOAP~\citep{vyas2024soap} may scale differently. In the context of LLMs, these optimizers seem to shift the intercept of the scaling law but do not affect the compute-optimal TPP~\citep{kimiteam2025kimik2}.
    \item \textbf{Text Encoder:} We only ever used a single text encoder in this work, however we expect that variations in the text encoder will have some measurable effect on the output quality of the models. This will necessarily shift the scaling laws in some way.
    \item \textbf{Vision Encoder:} In the context of LLMs the role of the text encoder in scaling behavior is an actively studied area~\citep{tao2024scaling,limisiewicz2026compute}. We did not ablate the vision encoder in our work but it seems likely that the choice of vision encoder will materially affect the scaling behavior of text-to-image generation.
    \item \textbf{Batch Size Scaling:} Compute-optimal batch size scaling studies have only recently been conducted in the much cheaper context of LLM scaling~\citep{bergsma2025power}. We suspect that there exist analogous compute-optimal batch size scaling laws for diffusion models but the empirical study of these scaling laws is beyond the scope of the present work.
\end{enumerate}

%% file: content/a03_heterogeneous.tex
\cutsectionup
\section{The Effect of Diffusion Timestep}\label{subsec:timestep}
\cutsectiondown

\begin{tcolorbox}
    \textbf{Finding:} Flow-matching loss is heterogeneous across diffusion timesteps, yet each timestep on its own continues to obey a scaling law.
\end{tcolorbox}

The training loss aggregates per-timestep losses weighted by the logit-normal timestep distribution. It is useful to isolate the per-sample loss and its expectation, the \textbf{loss response curve},
\begin{align*}
    \mathcal{L}(v_\theta, t; x, \varepsilon)&:=\frac{1}{n}\lVert v_\theta(t; x_t)-(\varepsilon - x)\rVert_{L^2}^2, \\
    \mathcal{L}^*(v_\theta, t)&:= \mathbb{E}_{\substack{x\in \Omega \\ \varepsilon \sim \mathcal{N}(0, \bm{I})}}\left[\,\mathcal{L}(v_\theta, t;x, \varepsilon)\,\right].
\end{align*}
\cutequationdown
We can then consider the oracle loss $v^*(t) = \inf_{v(t)\in C^1}\mathcal{L}^*(v; t)$, the minimizer of the loss response curve over an appropriate function space. This oracle is intractable on the interior, but its value at the two endpoints follows easily. At $t=0$ (a clean image) we have
\begin{align*}
    v^*(0; x) = -x, \mathcal{L}^*(v^*, 0) = \mathbb{E}_{\varepsilon\sim \mathcal{N}(0, 1)}\left[\lVert\varepsilon\rVert_{L^2}^2\right] = 1,
\end{align*}
\cutequationdown
and at $t=1$ (pure noise)
\begin{align*}
    v^*(1; x) = \varepsilon, \mathcal{L}^*(v^*, 1) = \mathbb{E}_{x\sim \Omega}\left[\lVert x\rVert_{L^2}^2\right],
\end{align*}
\cutequationdown
a constant fixed empirically by the data distribution $\Omega$. Both follow because $x$ and $\varepsilon$ are independent: at each endpoint the model receives no information about the other, and the loss is minimized by guessing that quantity's mean.

\begin{figure}[!ht]
    \centering
    \includegraphics[width=0.95\textwidth]{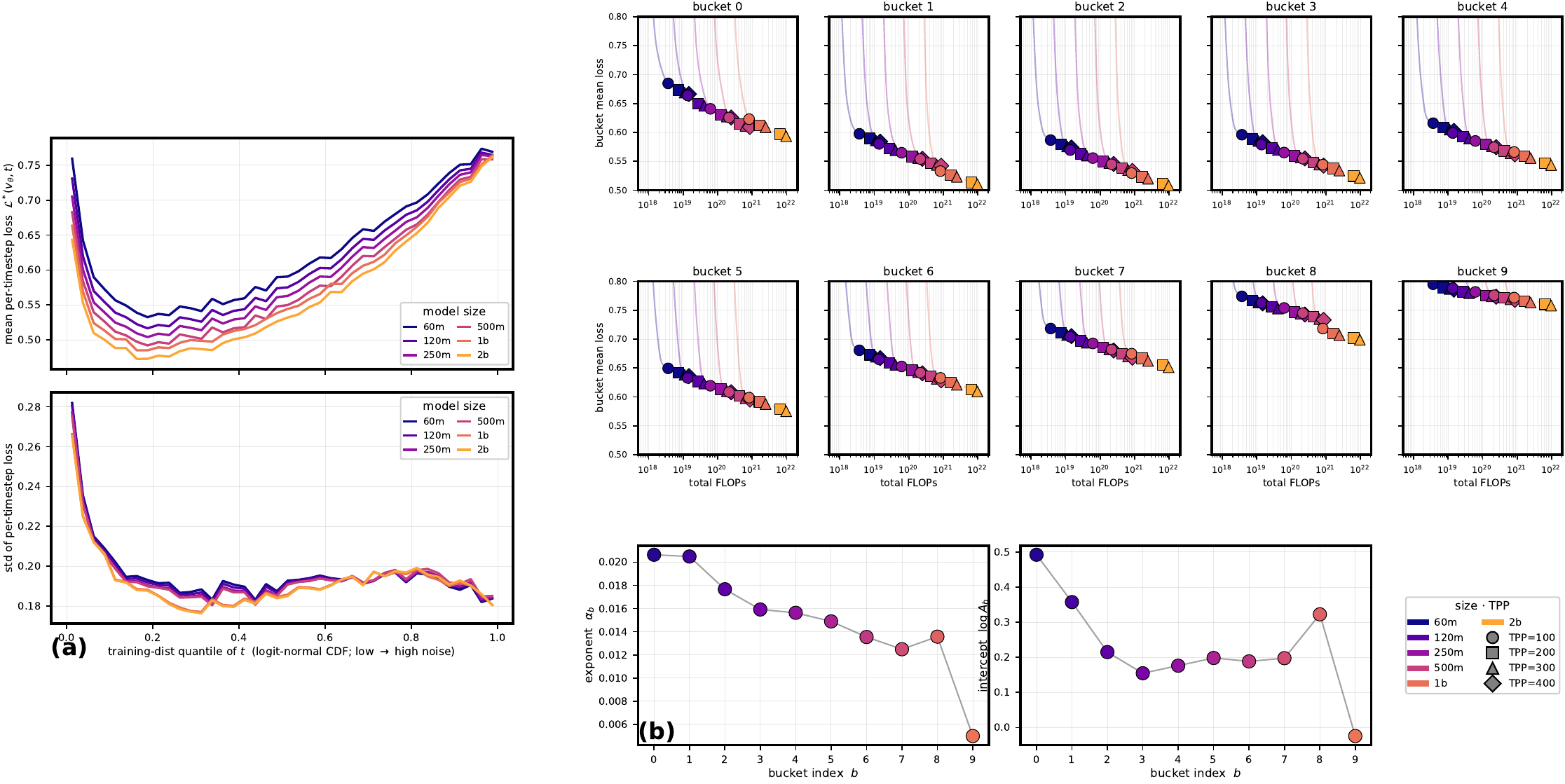}
    \cutcaptionup
    \caption{Timestep loss response and per-bucket scaling.
    \textbf{Left:} the per-timestep loss response curves $\mathcal{L}^*(v_\theta, t)$ (mean, top; std, bottom) at the final checkpoint for the \abra family. \textbf{Right:} each timestep bucket individually obeys a clean scaling law. \emph{Top two rows:} scaling laws in EMA model loss for 10 timestep buckets. \emph{Bottom:} the exponents $\alpha_b$ and biases $\log A_b$ of the supporting-hyperplane fits per bucket.}
    \cutcaptiondown
    \label{fig:bucketed_loss}
\end{figure}

In practice this loss response curve $\mathcal{L}^*(v_\theta, t)$ is highly heterogeneous. We measure it by sampling the EMA model 100k times using the logit-normal training distribution and bucketing the individual per-timestep losses (Figure~\ref{fig:bucketed_loss}). The endpoints sit near their oracle floors, while the interior timesteps carry the bulk of the reducible loss.

Despite the heterogeneity each timestep bucket individually obeys a clean scaling law. Binning the trajectory into ten uniform buckets and fitting the per-bucket loss law $L_b(C) = A_b\,C^{-\alpha_b}$ yields ten well-behaved fits (see Figure~\ref{fig:bucketed_loss}). The scaling exponents differ across buckets implying that the rate at which different timesteps reach compute optimality is heterogeneous: some timesteps saturate compute early while others continue to learn rapidly.

%% file: content/a04_model_generations.tex
\cutsectionup
\section{Model Generations}
\begin{figure}[!ht]
    \centering
    \ifshowgenerations
        \includegraphics[width=0.95\textwidth]{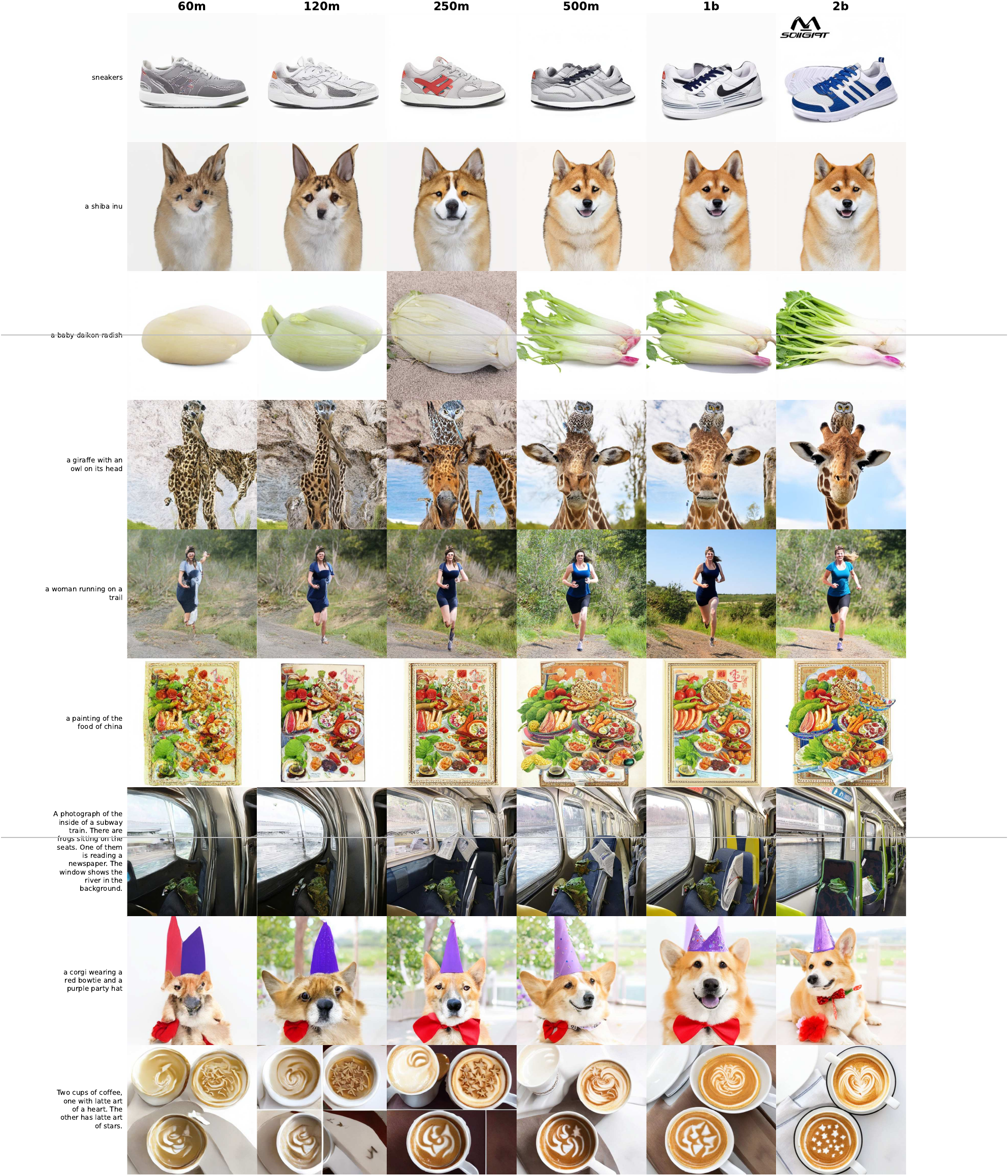}
    \else
        \omittedfig
    \fi
    \cutcaptionup
    \caption{Non-cherry-picked generations from the \abra family on Parti prompts~\citep{yu2022parti} at a fixed guidance scale of $3.0$. Columns are increasing model size (\abrasize{60M}$\to$\abrasize{2B}); rows are three prompts each of simple, standard, and complex complexity (word-count terciles). Each cell is a single $512\times512$ generation.}
    \cutcaptiondown
    \label{fig:non_cherry_picked_generations}
\end{figure}

\begin{figure}[!ht]
    \centering
    \ifshowgenerations
        \includegraphics[width=0.95\textwidth]{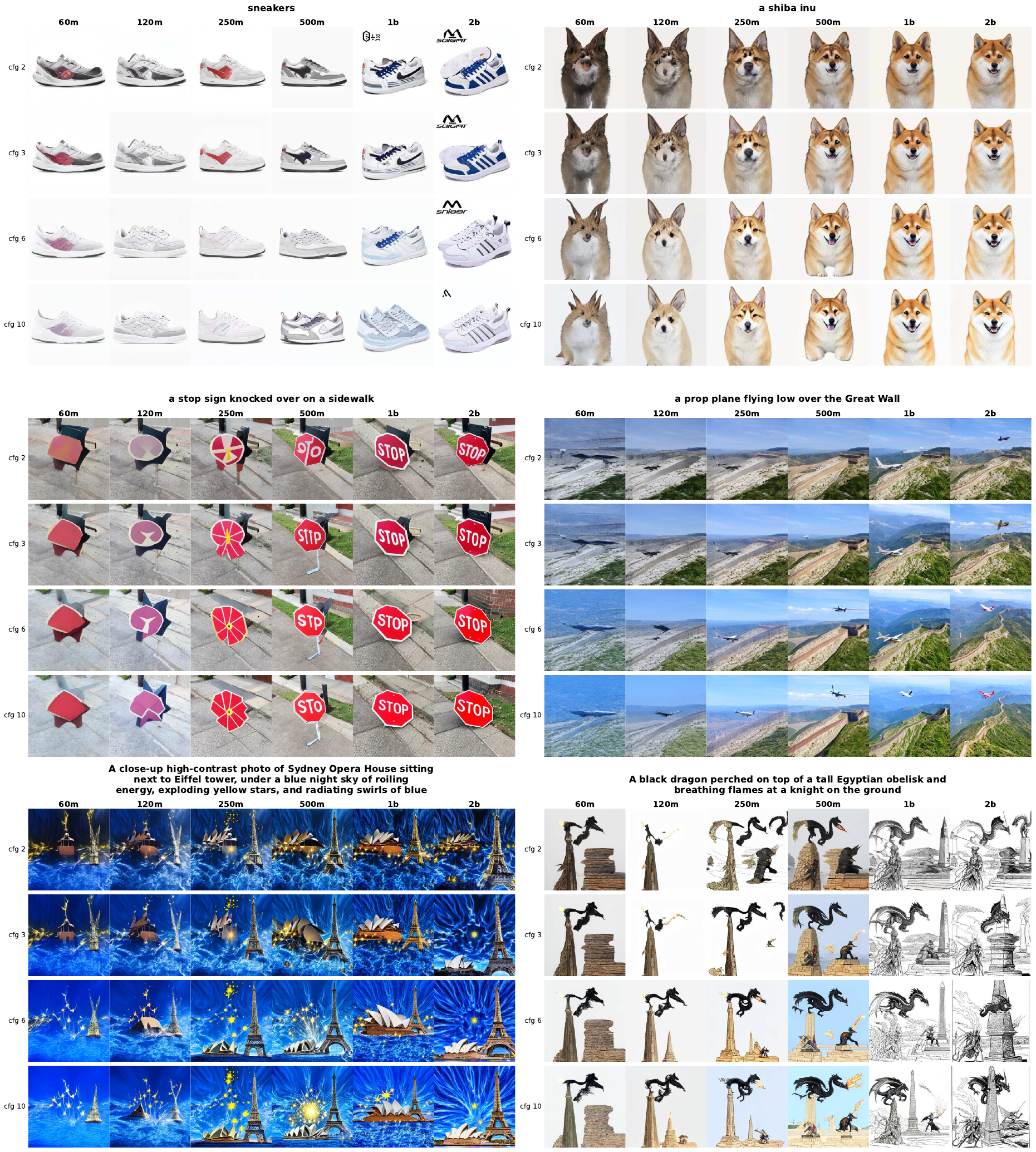}
    \else
        \omittedfig
    \fi
    \cutcaptionup
    \caption{Model size vs. CFG at the compute-optimal ($200$-TPP) checkpoint. Panels are two Parti prompts~\citep{yu2022parti} each of simple, standard, and complex complexity (rows). Within each panel, columns are increasing model size (\abrasize{60M}$\to$\abrasize{2B}) and rows are the guidance scale (CFG $\in\{2, 3, 6, 10\}$). Each cell is a single $512\times512$ generation.}
    \cutcaptiondown
    \label{fig:size_cfg}
\end{figure}

\cutsectiondown
We share non-cherry-picked sample generations using prompts from the Parti Prompts dataset~\citep{yu2022parti}. We sorted the prompts into ``simple'', ``standard'', and ``complex'' based on the length of the prompt. Figure~\ref{fig:non_cherry_picked_generations} shows the generated responses of our models at a fixed guidance scale of $3.0$. Figure~\ref{fig:size_cfg} shows the generations for each model size with varying guidance scale.

%% file: content/a05_extended_scaling_analysis.tex
\cutsectionup
\section{Extended Scaling Analysis}
\cutsectiondown
\label{app:extended_scaling}

\begin{figure}[!ht]
    \centering
    \includegraphics[width=0.95\textwidth]{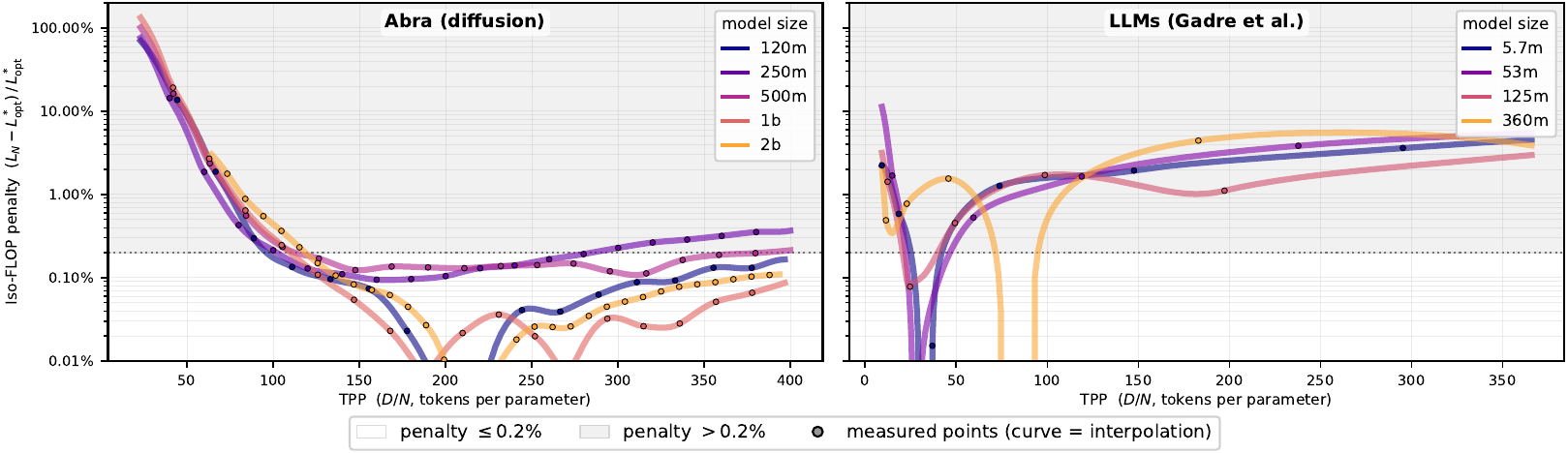}
    \cutcaptionup
    \caption{Iso-FLOP relative loss penalty $\Delta L / L^*$ versus TPP for
    \emph{(left)} the \abra family and \emph{(right)} the overtrained LLM grid of
   ~\citet{gadre2025language}. Markers denote the measured points (\abra
    evaluated checkpoints; Gadre et al.'s per-run final losses); the connecting
    curves interpolate through them. Doubling the optimal TPP for the largest
    model incurs a ${\sim}4\%$ loss penalty --- nearly $20\times$ that of the
    diffusion model.}
    \cutcaptiondown
    \label{fig:gadre_penalty}
\end{figure}

\begin{figure}[!ht]
    \centering
    \includegraphics[width=0.95\textwidth]{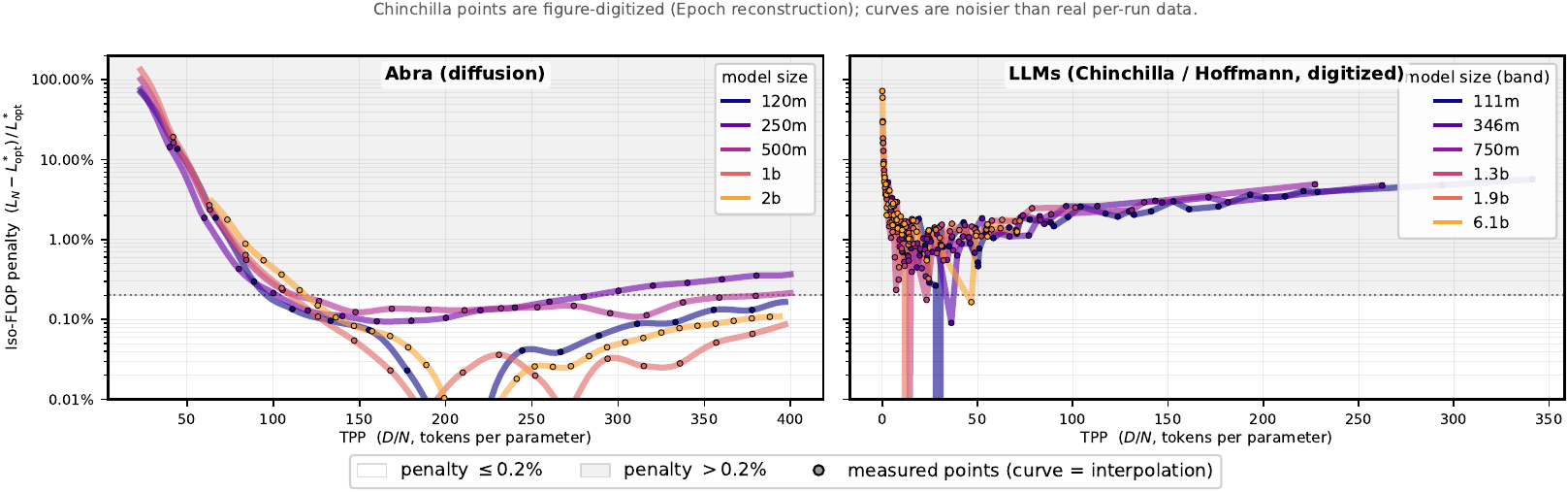}
    \cutcaptionup
    \caption{Iso-FLOP suboptimality penalty $(L_N - L^*)/L^*$ versus TPP for the
    \abra family \emph{(left)} and the Chinchilla / Hoffmann et al.\ data
    \emph{(right)}, the latter figure-digitized by~\citet{besiroglu2024chinchilla}
    and binned into log-$N$ model-size bands. Markers are the measured points
    (\abra evaluated checkpoints; digitized Chinchilla points); curves interpolate
    through them. The green/red bands mark the $0.2\%$ penalty threshold.}
    \cutcaptiondown
    \label{fig:penalty_chinchilla}
\end{figure}

\begin{figure}[!ht]
    \centering
    \includegraphics[width=0.95\textwidth]{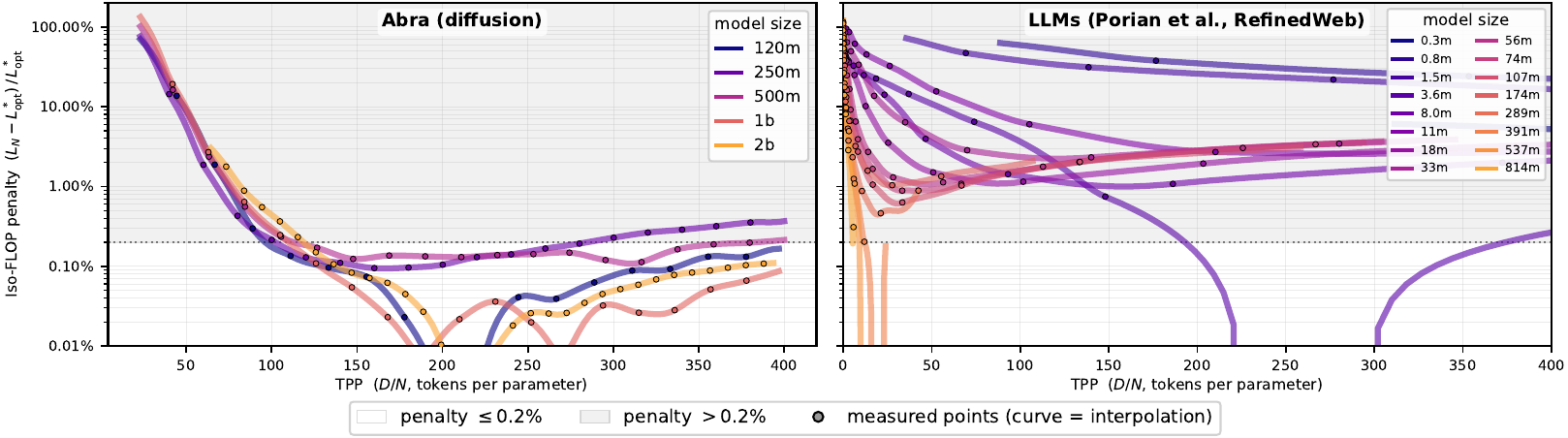}
    \cutcaptionup
    \caption{Iso-FLOP suboptimality penalty $(L_N - L^*)/L^*$ versus TPP for the
    \abra family \emph{(left)} and the LLM runs of~\citet{porian2024resolving}
    on RefinedWeb \emph{(right)}. Markers are the measured points (\abra evaluated
    checkpoints; Porian et al.'s per-step logged validation losses along each
    size's longest run); curves interpolate through them. The green/red bands mark
    the $0.2\%$ penalty threshold.}
    \cutcaptiondown
    \label{fig:penalty_porian}
\end{figure}

In Section~\ref{subsec:overtraining} above we compared the iso-FLOP suboptimality penalty of the \abra family of models to the LLM family trained by~\citet{mcleish2025gemstones} (see Figure~\ref{fig:avg_loss_penalty}). In this section we include the same analysis carried out against other open-source LLM pre-training data. In particular, we compare against~\citet{gadre2025language} (see Figure~\ref{fig:gadre_penalty}),~\citet{porian2024resolving} (Figure~\ref{fig:penalty_porian}), and the Chinchilla paper~\citep{hoffmann2022empirical} (Figure~\ref{fig:penalty_chinchilla}). We are not aware of an openly available source for the Chinchilla study's data and thus use digitization to extract loss curves from the paper's figures.

%% file: content/a06_fit_procedures.tex
\cutsectionup
\section{Fit Procedures, Alternatives, and the Kaplan/Hoffmann Recipes}
\cutsectiondown
\label{app:fit_procedures}
We detail our fitting methodology, with error and quality-of-fit analysis, reproduce the two canonical language-model recipes of~\citet{kaplan2020scaling} and~\citet{hoffmann2022empirical}, and test several alternative estimators. All fits are collected in Table~\ref{tab:procedure_comparison}.

\label{app:our_procedure}
The main-body procedure (Section~\ref{subsec:tpp_estimation}) fits per-TPP power
laws $L(C;\mathrm{TPP}) = A\,C^{-\alpha} + \beta$ to the EMA-weight loss and reads
the optimum off iso-FLOP slices, interpolating each trajectory with
PCHIP~\citep{fritsch1980monotone, fritsch1984method} and counting $N$ as all
diffusion-transformer parameters (frozen text encoder excluded). The per-TPP fits
are trust-region non-linear least squares on the raw loss values in the physical
space, after discarding the first $5{,}000$ warmup steps.

\cutsubsectionup
\subsection{Supporting-Hyperplane Fit}
\cutsubsectiondown
\label{app:hyperplane}
The compute-optimal frontier is a maximal lower bound of the family's loss curves,
so we fit it directly as the maximal power-law lower bound of the measured
$(N, C, L)$ points. Over a grid of
exponents $\alpha$ we solve for the amplitude and offset $(A, \beta)$ maximizing the
fitted loss subject to $\hat{L}_i := A\,C_i^{-\alpha} + \beta \le L_i$ at every
Pareto point $i$, then select the $\alpha$ of minimal total slack
$\sum_i (L_i - \hat{L}_i)$. On \abra{} this recovers the same $L^*(C)$ as the
iso-FLOP envelope to within $<\!0.1\%$ (coefficients in
Table~\ref{tab:procedure_comparison_fits}).

\cutsubsectionup
\subsection{Kaplan-Style Fit}
\cutsubsectiondown
\label{app:kaplan_fit}

\citet{kaplan2020scaling} fit a pure power law $L(C) \propto C^{-\alpha_C}$ to the
compute-efficient lower envelope; unlike the supporting hyperplane of
Section~\ref{app:hyperplane}, it regresses a curve \emph{through} the envelope
points rather than a bound weakly below them. We fit the allocation exponents both
freely and constrained to $a + b = 1$ (Table~\ref{tab:kh_allocation}).

\begin{table}[!ht]
    \centering
    \cutcaptionup
    \caption{Compute-optimal allocation exponents ($N_{\text{opt}} \propto C^{a}$,
    $D_{\text{opt}} \propto C^{b}$) for the \abra envelope. The \emph{free} fit
    regresses $\log N_{\text{opt}}$ and $\log D_{\text{opt}}$ on $\log C$
    independently; the \emph{constrained} fit imposes $a+b=1$. The free $a+b>1$
    reflects the per-model transformer FLOP count exceeding the $6ND$ estimate by
    a size-dependent factor; the balanced split is robust either way.}
    \cutcaptiondown
    \label{tab:kh_allocation}
    \input{tables/kh_allocation}
\end{table}

\cutsubsectionup
\subsection{Hoffmann-Style (Chinchilla) Fits}
\cutsubsectiondown
\label{app:hoffmann_fit}
\citet{hoffmann2022empirical} estimate the compute-optimal frontier three
distinct ways, and we reproduce the applicable ones for the \abra family.

\paragraph{Approach 1: Minima over training curves.}
For each FLOP budget $C$ we read off the model attaining the lowest loss at that
budget, giving $N_{\text{opt}}(C)$ and $D_{\text{opt}}(C)$ directly from the
envelope, and take the median over budgets as the implied
$\mathrm{TPP}^*$ (Table~\ref{tab:procedure_comparison}).

\paragraph{Approach 2: Iso-FLOP profiles.}
This is the estimator we use in the main body (Section~\ref{subsec:tpp_estimation}).
For each of a grid of fixed FLOP budgets spanning the compute range where the
six-model ladder overlaps, we take the loss-vs-TPP slice of the fitted surface and
read off its minimum, then aggregate over budgets (Table~\ref{tab:procedure_comparison}).

\paragraph{Approach 3: Parametric loss surface.}
One can fit the parametric form
\begin{align*}
    L(N, D) = E + \frac{A}{N^{\alpha}} + \frac{B}{D^{\beta}}
\end{align*}
\cutequationdown
to the EMA loss in log space with a Huber objective ($\delta = 10^{-3}$) and a grid
of L-BFGS initializations, following~\citet{hoffmann2022empirical}, then derive
$N_{\text{opt}}(C)$ and $D_{\text{opt}}(C)$ from the fitted surface. On the \abra
ladder this fit is non-identifiable and we report no parametric $\mathrm{TPP}^*$ fit.

\cutsubsectionup
\subsection{Alternative Procedures and Unified Comparison}
\cutsubsectiondown
\label{app:alternatives}
\label{app:procedure_comparison}
Beyond the estimator, we re-estimate the compute-optimal TPP under one-at-a-time
variations of the pipeline:
\begin{itemize}
    \item \textbf{Fit target.} EMA loss (headline) vs.\ raw training loss.
    \item \textbf{Ladder composition.} Full ladder vs.\ dropping the smallest
    (\abrasize{60M}) and/or largest (\abrasize{2B}) models.
    \item \textbf{Fit objective.} Squared error (headline) vs.\ Huber.
    \item \textbf{Interpolation.} PCHIP (headline) vs.\ piecewise-linear
    interpolation directly between checkpoints, with no shape-preserving smoothing.
\end{itemize}
Table~\ref{tab:procedure_comparison} collects the recovered $\mathrm{TPP}^*$ from
every estimator and variation, with 60M included and excluded; the corresponding
frontier and allocation coefficients are in
Table~\ref{tab:procedure_comparison_fits}.

\begin{table}[!ht]
    \centering
    \cutcaptionup
    \caption{Compute-optimal $\mathrm{TPP}^*$ under every estimator and procedure
    variation on identical \abra data, with 60M included and excluded (holding the
    other choice fixed); the headline (iso-FLOP, 60M excluded) is \textbf{bold}.}
    \cutcaptiondown
    \label{tab:procedure_comparison}
    \small
    \setlength{\tabcolsep}{4pt}
    \begin{tabular}{lccp{3.9cm}}
        \toprule
        Estimator / variation & TPP* (incl.\ 60M) & TPP* (excl.\ 60M) & Notes \\
        \midrule
        Iso-FLOP (main body)         & $186$ & $\mathbf{199}$ & headline; $68\%$ eval-noise $[180,270]$ \\
        Kaplan (envelope OLS)        & $185$ & $192$ & balanced allocation $a\approx0.47$ \\
        Supporting hyperplane (LP)   & $185$ & $192$ & same envelope allocation as Kaplan \\
        Hoffmann A1 (curve minima)   & $185$ & $192$ & noisy minima; wide p10--p90 \\
        Raw-loss target              & $189$ & $227$ & only material mover; not used \\
        Huber fit objective          & $186$ & $199$ & identical to least squares \\
        Linear interp.\ (vs.\ PCHIP) & $187$ & $202$ & piecewise-linear between checkpoints \\
        Drop \abrasize{2B}           & $183$ & $195$ & ladder-endpoint check \\
        \bottomrule
    \end{tabular}
\end{table}

\begin{table}[!ht]
    \centering
    \scriptsize
    \setlength{\tabcolsep}{4.5pt}
    \cutcaptionup
    \caption{Full fit-coefficient comparison, the companion to
    Table~\ref{tab:procedure_comparison} (which reports only the recovered
    $\mathrm{TPP}^*$). For each of the seven estimators and variations we report the
    compute-optimal loss frontier $L^*(C) = A\,C^{-\alpha} + \beta$ together with
    the optimal-allocation laws $N_{\text{opt}}(C) = A_N\,C^{a}$ and
    $D_{\text{opt}}(C) = A_D\,C^{b}$, each computed with the 60M model included and
    excluded, and under three allocation conventions: \emph{free} (independent
    log--log fits, $a{+}b \approx 1.04$), constrained ($a{+}b{=}1$), and balanced
    ($a{=}b{=}0.5$).}
    \cutcaptiondown
    \label{tab:procedure_comparison_fits}
    \begin{tabular}{lll ccc cccc}
        \toprule
        Method & Ladder & Alloc. & $A$ & $\alpha$ & $\beta$ & $A_N$ & $a$ & $A_D$ & $b$ \\
        \midrule
        Iso-FLOP & incl.\ 60M & free & $1.37$ & $0.0371$ & $0.3874$ & $0.0211$ & $0.503$ & $0.978$ & $0.533$ \\
         &  & $a{+}b{=}1$ &  &  &  & $0.049$ & $0.485$ & $2.27$ & $0.515$ \\
         &  & $a{=}b{=}0.5$ &  &  &  & $0.0243$ & $0.500$ & $4.57$ & $0.500$ \\
        \cmidrule(l){2-10}
         & excl.\ 60M & free & $1.2$ & $0.0140$ & $0.0052$ & $0.0126$ & $0.514$ & $1.58$ & $0.523$ \\
         &  & $a{+}b{=}1$ &  &  &  & $0.0301$ & $0.495$ & $3.78$ & $0.505$ \\
         &  & $a{=}b{=}0.5$ &  &  &  & $0.0239$ & $0.500$ & $4.75$ & $0.500$ \\
        \midrule
        Supporting hyperplane & incl.\ 60M & free & $1.19$ & $0.0140$ & $0.0077$ & $0.051$ & $0.484$ & $0.431$ & $0.550$ \\
         &  & $a{+}b{=}1$ &  &  &  & $0.115$ & $0.467$ & $0.97$ & $0.533$ \\
         &  & $a{=}b{=}0.5$ &  &  &  & $0.0245$ & $0.500$ & $4.54$ & $0.500$ \\
        \cmidrule(l){2-10}
         & excl.\ 60M & free & $1.19$ & $0.0140$ & $0.0077$ & $0.0444$ & $0.487$ & $0.491$ & $0.548$ \\
         &  & $a{+}b{=}1$ &  &  &  & $0.102$ & $0.470$ & $1.12$ & $0.530$ \\
         &  & $a{=}b{=}0.5$ &  &  &  & $0.0244$ & $0.500$ & $4.69$ & $0.500$ \\
        \midrule
        Hoffmann A1 & incl.\ 60M & free & $1.42$ & $0.0386$ & $0.3965$ & $0.051$ & $0.484$ & $0.431$ & $0.550$ \\
         &  & $a{+}b{=}1$ &  &  &  & $0.115$ & $0.467$ & $0.97$ & $0.533$ \\
         &  & $a{=}b{=}0.5$ &  &  &  & $0.0245$ & $0.500$ & $4.54$ & $0.500$ \\
        \cmidrule(l){2-10}
         & excl.\ 60M & free & $1.21$ & $0.0139$ & $0$ & $0.046$ & $0.487$ & $0.509$ & $0.547$ \\
         &  & $a{+}b{=}1$ &  &  &  & $0.102$ & $0.470$ & $1.12$ & $0.530$ \\
         &  & $a{=}b{=}0.5$ &  &  &  & $0.0244$ & $0.500$ & $4.69$ & $0.500$ \\
        \midrule
        Huber & incl.\ 60M & free & $1.37$ & $0.0371$ & $0.3873$ & $0.0211$ & $0.503$ & $0.978$ & $0.533$ \\
         &  & $a{+}b{=}1$ &  &  &  & $0.049$ & $0.485$ & $2.27$ & $0.515$ \\
         &  & $a{=}b{=}0.5$ &  &  &  & $0.0243$ & $0.500$ & $4.57$ & $0.500$ \\
        \cmidrule(l){2-10}
         & excl.\ 60M & free & $1.2$ & $0.0140$ & $0.0090$ & $0.0126$ & $0.514$ & $1.58$ & $0.523$ \\
         &  & $a{+}b{=}1$ &  &  &  & $0.0302$ & $0.495$ & $3.77$ & $0.505$ \\
         &  & $a{=}b{=}0.5$ &  &  &  & $0.0239$ & $0.500$ & $4.75$ & $0.500$ \\
        \midrule
        Linear interp & incl.\ 60M & free & $1.37$ & $0.0370$ & $0.3864$ & $0.0211$ & $0.503$ & $0.978$ & $0.533$ \\
         &  & $a{+}b{=}1$ &  &  &  & $0.049$ & $0.485$ & $2.27$ & $0.515$ \\
         &  & $a{=}b{=}0.5$ &  &  &  & $0.0243$ & $0.500$ & $4.57$ & $0.500$ \\
        \cmidrule(l){2-10}
         & excl.\ 60M & free & $1.2$ & $0.0140$ & $0.0076$ & $0.0123$ & $0.514$ & $1.61$ & $0.523$ \\
         &  & $a{+}b{=}1$ &  &  &  & $0.0295$ & $0.495$ & $3.86$ & $0.505$ \\
         &  & $a{=}b{=}0.5$ &  &  &  & $0.0238$ & $0.500$ & $4.79$ & $0.500$ \\
        \midrule
        Raw-loss & incl.\ 60M & free & $1.13$ & $0.0159$ & $0$ & $1.11$ & $0.419$ & $0.0246$ & $0.611$ \\
         &  & $a{+}b{=}1$ &  &  &  & $2.25$ & $0.404$ & $0.0497$ & $0.596$ \\
         &  & $a{=}b{=}0.5$ &  &  &  & $0.0254$ & $0.500$ & $4.4$ & $0.500$ \\
        \cmidrule(l){2-10}
         & excl.\ 60M & free & $1.14$ & $0.0162$ & $0.0020$ & $0.00152$ & $0.558$ & $11.3$ & $0.482$ \\
         &  & $a{+}b{=}1$ &  &  &  & $0.00391$ & $0.538$ & $29$ & $0.462$ \\
         &  & $a{=}b{=}0.5$ &  &  &  & $0.0234$ & $0.500$ & $4.85$ & $0.500$ \\
        \midrule
        Drop 2B & incl.\ 60M & free & $1.97$ & $0.0528$ & $0.4623$ & $0.024$ & $0.500$ & $0.868$ & $0.535$ \\
         &  & $a{+}b{=}1$ &  &  &  & $0.0548$ & $0.482$ & $1.98$ & $0.518$ \\
         &  & $a{=}b{=}0.5$ &  &  &  & $0.0245$ & $0.500$ & $4.44$ & $0.500$ \\
        \cmidrule(l){2-10}
         & excl.\ 60M & free & $1.19$ & $0.0141$ & $0.0153$ & $0.00533$ & $0.532$ & $3.51$ & $0.506$ \\
         &  & $a{+}b{=}1$ &  &  &  & $0.013$ & $0.513$ & $8.55$ & $0.487$ \\
         &  & $a{=}b{=}0.5$ &  &  &  & $0.0239$ & $0.500$ & $4.65$ & $0.500$ \\
        \bottomrule
    \end{tabular}
\end{table}

\cutsubsectionup
\subsection{Choice of Interpolant}
\cutsubsectiondown
\label{app:interpolation_error}
We interpolate each model's EMA-loss-versus-step trajectory with a shape-preserving
PCHIP interpolant~\citep{fritsch1980monotone, fritsch1984method}, and compare against
plain piecewise-linear interpolation.

\paragraph{Leave-one-out interpolation error.}
For each model we hold out one interior checkpoint, refit on the remaining points,
and predict the held-out loss; Table~\ref{tab:loo_rmse} reports the RMSE over all
held-out interior points.

\begin{table}[!ht]
    \centering
    \cutcaptionup
    \caption{Leave-one-out interpolation RMSE per model, for PCHIP and
    piecewise-linear interpolation; PCHIP is the more accurate and both are small.}
    \cutcaptiondown
    \label{tab:loo_rmse}
    \begin{tabular}{lccccc}
        \toprule
        Interpolant & \abrasize{60M} & \abrasize{120M} & \abrasize{250M} & \abrasize{500M} & \abrasize{1B} \\
        \midrule
        PCHIP (main body) & $0.0167$ & $0.0171$ & $0.0233$ & $0.0317$ & $0.0430$ \\
        Piecewise-linear  & $0.0379$ & $0.0406$ & $0.0513$ & $0.0672$ & $0.0882$ \\
        \bottomrule
    \end{tabular}
\end{table}

\cutsubsectionup
\subsection{Leave-One-Model-Out Robustness}
\cutsubsectiondown
\label{app:lomo}
To bound the leverage of any single model we refit the iso-FLOP $\mathrm{TPP}^*$
holding out each model in turn, for both ladder choices (Table~\ref{tab:lomo}). With the exception of \abrasize{60M}, leaving out a single model does not materially affect the fit.

\begin{table}[!ht]
    \centering
    \cutcaptionup
    \caption{Leave-one-model-out iso-FLOP $\mathrm{TPP}^*$: the estimate with each
    model held out of the ladder, for the 60M-included and 60M-excluded bases. With 60M excluded, every
    held-out estimate stays in $[188, 207]$.}
    \cutcaptiondown
    \label{tab:lomo}
    \begin{tabular}{lcc}
        \toprule
        Held out & TPP* (incl.\ 60M base) & TPP* (excl.\ 60M base) \\
        \midrule
        none (full)     & $186$ & $199$ \\
        \abrasize{60M}  & $199$ & --- \\
        \abrasize{120M} & $170$ & $188$ \\
        \abrasize{250M} & $189$ & $207$ \\
        \abrasize{500M} & $185$ & $198$ \\
        \abrasize{1B}   & $184$ & $203$ \\
        \abrasize{2B}   & $183$ & $195$ \\
        \bottomrule
    \end{tabular}
\end{table}

\cutsubsectionup
\subsection{Every Estimator Agrees}
\cutsubsectiondown
The headline ${\approx}200$ TPP and the ${\sim}10\times$ gap compared to the
LLM value of $20$ is method-invariant. Every identifiable estimator lands in
$[183, 202]$ (Table~\ref{tab:procedure_comparison}) and the balanced allocation
($a \approx 0.47$--$0.49$) holds under both free and constrained fits; the estimate
is likewise unchanged under either interpolant (Section~\ref{app:interpolation_error})
and under holding out any single model (Section~\ref{app:lomo}).

\cutsubsectionup
\subsection{Generative Metric Scaling-Law Fits}\label{app:metric_fits}
\cutsubsectiondown

\begin{figure}[!ht]
    \centering
    \includegraphics[width=\textwidth]{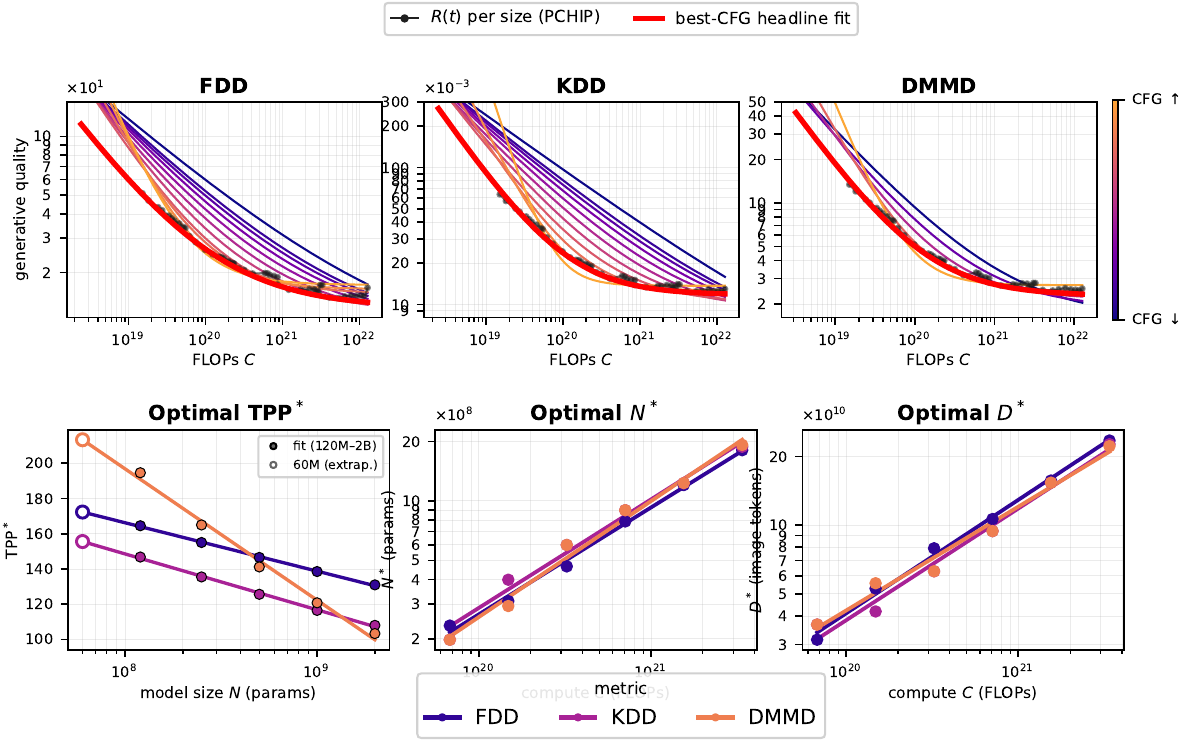}
    \cutcaptionup
    \caption{\textbf{DINOv2-family generative metrics scale predictably.} \emph{Top:} generation quality vs training compute $C$ for the three DINOv2-based distances---FDD (Fr\'echet), KDD (kernel), and DMMD (MMD). At each guidance scale the evaluation trajectories are fit with the iso-FLOP compute-optimal procedure of Section~\ref{subsec:tpp_estimation} (curves colored by CFG); dark markers and curves are the best-over-CFG trajectory $R(t)$ per model size, and the bold red curve is its compute-optimal frontier. \emph{Bottom:} the resulting compute-optimal $\mathrm{TPP}^*$ as a function of model size $N$ (linear fit through 120M--2B, 60M extrapolated), and the compute-optimal $N_{\text{opt}}$ and $D_{\text{opt}}$ as functions of compute $C$. All three DINOv2 distances reach compute optimality at a substantially lower TPP than the loss.}
    \cutcaptiondown
    \label{fig:generative_metrics_dino}
\end{figure}

This appendix details the compute-optimal $\mathrm{TPP}^*$ procedure for the generative metrics (Section~\ref{sec:generative_metrics},
Table~\ref{tab:metric_fits}) and reports its stability, then cross-checks the
metric-value scaling with a per-model-final fit. Figure~\ref{fig:generative_metrics_dino}
presents the full per-metric iso-FLOP fits for the three DINOv2-family distances
(FDD, KDD, and DMMD).

\begin{table}[!ht]
    \centering
    \cutcaptionup
    \caption{Compute-optimal tokens per parameter $\mathrm{TPP}^*$ per generative
    metric, recovered with the headline iso-FLOP procedure
    (Section~\ref{subsec:tpp_estimation}) applied to each metric's best-over-CFG
    trajectory.}
    \cutcaptiondown
    \label{tab:metric_fits}
    \input{tables/metric_fits}
\end{table}

\begin{figure}[!ht]
    \centering
    \includegraphics[width=0.95\textwidth]{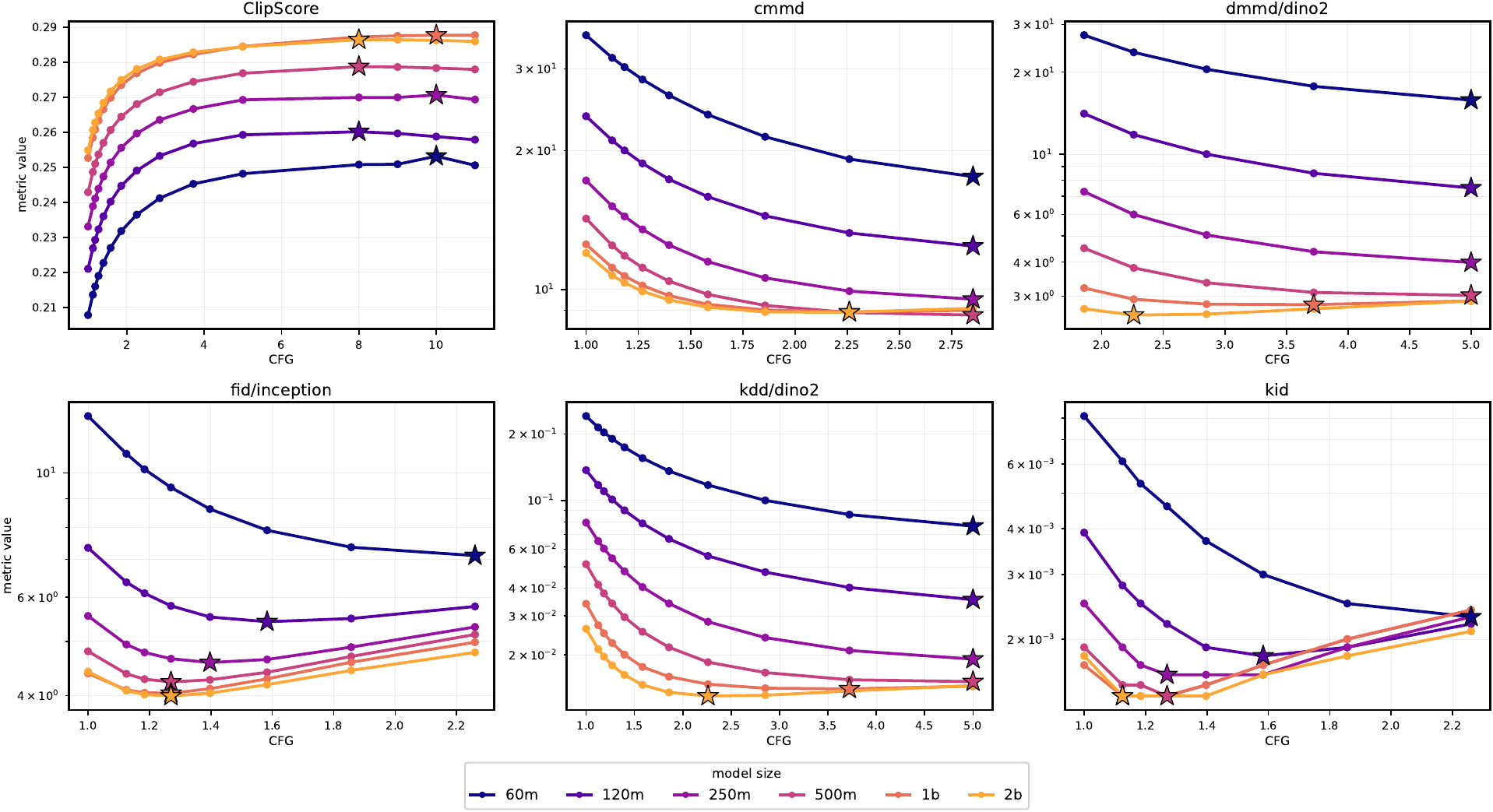}
    \cutcaptionup
    \caption{Each generative metric as a function of the classifier-free guidance
    scale, swept per model size. The optimum (a minimum for the distribution
    distances, a maximum for CLIPScore) shifts with model size, so comparing models
    at a single fixed guidance value would misrank them; this motivates the
    per-model CFG optimization we use throughout.}
    \cutcaptiondown
    \label{fig:cfg_sweeps}
\end{figure}

\begin{figure}[!ht]
    \centering
    \includegraphics[width=0.95\textwidth]{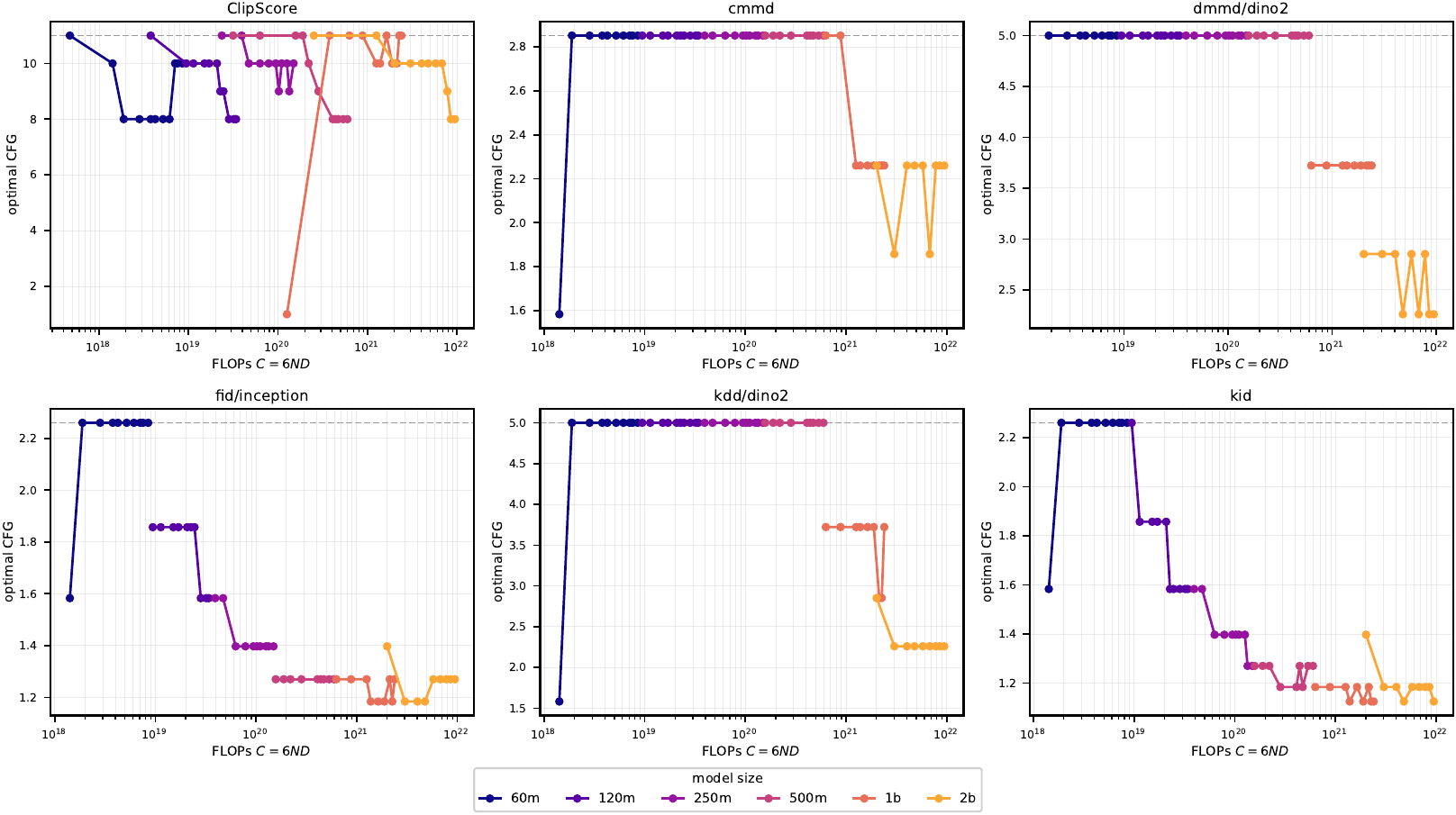}
    \cutcaptionup
    \caption{Optimal CFG versus training compute $C = 6ND$, one panel per metric,
    with each model size a different colour and every swept checkpoint plotted
    (not just the final one). The optimal CFG drifts downward with both model
    size and training compute for the distribution metrics.}
    \cutcaptiondown
    \label{fig:optimal_cfg_flops}
\end{figure}

\paragraph{Iso-FLOP $\mathrm{TPP}^*$ (primary).} We reuse the headline loss
estimator (Section~\ref{subsec:tpp_estimation}) verbatim, with the CFG-optimized
metric in place of the loss (min over the CFG grid for the distance metrics and
FID, max for CLIPScore). Per size we PCHIP-spline the metric against step; per TPP
on a grid ($80$--$390$, step $5$) we read each size at the step realizing that TPP,
convert to compute $C = \text{step}\cdot\text{GBS}\cdot F_N$, and fit
$M(C)=A\,C^{-\alpha}+F_M$ across the ladder excluding 60M ($\ge 3$ sizes). The
parabolic argmin over TPP of these per-TPP laws on six shared budgets gives
$\mathrm{TPP}^*(C)$; we report its geometric mean and log-log slope $p$
(Table~\ref{tab:metric_fits}). $\mathrm{TPP}^*$ differs markedly between metrics.

\paragraph{Value scaling law.} We fit the value law $M(C)=A\,C^{-\alpha}+F_M$
(sign flipped for CLIPScore) for two targets: the compute-optimal fit through the
$\mathrm{TPP}^*$ points and the per-model-final fit through each size's converged
checkpoint. We exclude 60M as elsewhere. Where the floor $F_M$ is not identifiable
(KDD, KID) we fit a pure power law $A\,C^{-\alpha}$; KID then gives
$\alpha\approx0.11$ ($R^2=0.90$). We fit each size rather than pooling checkpoints,
for which $M(C)$ is not single-valued ($R^2<0.1$ for the distribution metrics). The
exponents agree for DMMD, KDD, FID, and KID, and differ for CMMD and CLIPScore,
which improve further between $\mathrm{TPP}^*$ and convergence.

\paragraph{Dependence of the fit on CFG.} The value laws use the CFG-optimized
value per checkpoint. To check this selection is not the source of the scaling, we
refit the per-model-final law at every measured CFG and track $\alpha$ (dropping
fits with $R^2<0.5$ or $\alpha$ pinned at the optimizer bound). $\alpha$ is stable
across the swept range for every metric (Figure~\ref{fig:cfg_sweeps}): a mild
decline at high guidance for CMMD and CLIPScore, flat for the DINOv2 distances
DMMD/KDD. The scaling laws are therefore robust to CFG; the compute-optimal fit in
the body represents the fixed-CFG family rather than a special case.

%% file: tables/kh_allocation.tex
\begin{tabular}{lcccc}
    \toprule
    Ladder & $a$ (constr.) & $a$ (free) & $b$ (free) & $a+b$ (free) \\
    \midrule
        incl.\ 60M & $0.467$ & $0.484$ & $0.550$ & $1.035$ \\
        excl.\ 60M & $0.470$ & $0.487$ & $0.547$ & $1.033$ \\
    \bottomrule
\end{tabular}

%% file: tables/metric_fits.tex
\begin{tabular}{llcc}
    \toprule
    Metric & Backbone & $\mathrm{TPP}^*$ & $\mathrm{TPP}^*(C)$ exponent $p$ \\
    \midrule
    KDD       & DINOv2    & $121$ & $-0.06$ \\
    DMMD      & DINOv2    & $133$ & $-0.13$ \\
    CLIPScore & CLIP      & $153$ & $-0.05$ \\
    FID       & Inception & $182$ & $+0.12$ \\
    KID       & Inception & $257$ & $+0.12$ \\
    CMMD      & CLIP      & $294$ & $-0.14$ \\
    \midrule
    \emph{Loss (ref.)} & bucketed & $\approx\!200$ & $\approx 0$ \\
    \bottomrule
\end{tabular}

%% file: content/a07_parameter_counting.tex
\cutsectionup
\section{Comparison to~\citet{liang2024scaling}}
\cutsectiondown
\label{app:n_convention}

Table~\ref{tab:compare} compares our compute-optimal allocation law to that of
\citet{liang2024scaling}. Because~\citet{liang2024scaling} measure data in
\emph{total} (text-plus-image) tokens whereas we measure image tokens, only the
allocation \emph{exponents} are directly comparable across the two studies; the
multipliers and intercepts are not.

Note that the text encoding adds a certain number of tokens per sample which can be averaged over and will only serve to increase the estimated TPP. When using $256$ text tokens per sample, an image TPP of $200$ at $512\times 512$ resolution gives us a total TPP of $250$. For a 1B model,~\citet{liang2024scaling} find that optimal TPP should occur at $287$ total TPP, counting both text and image tokens. Note that~\citet{liang2024scaling} only report optimal $N$ and $D$ for $256\times 256$ images, which based on our analysis should have a lower optimal TPP than the $512\times 512$ images. Thus, we expect that $287$ over-estimates the number of training tokens which is optimal for training a 1B model.

\begin{table}[!ht]
    \centering
    \cutcaptionup
    \caption{Compute-optimal allocation laws for \abra,
    \abrasize{60M} excluded, versus~\citet{liang2024scaling}. $C$ in FLOPs, $N$ in parameters, $D$ in image tokens; both fits impose $a+b=1$. Note that our study measures compute in image tokens while~\citet{liang2024scaling} measure in total tokens. While this makes the intercepts incomparable, the scaling exponents \textbf{are} comparable between the two studies.}
    \cutcaptiondown
    \label{tab:compare}
    \input{tables/compare}
\end{table}

%% file: tables/compare.tex
\begin{tabular}{lcc}
    \toprule
    Allocation law & Ours (\abra) & \citet{liang2024scaling} \\
    \midrule
    $N_{\text{opt}}(C)$ & $0.0301\,C^{0.4951}$ & $0.0009\,C^{0.5681}$ \\
    $D_{\text{opt}}(C)$ & $3.776\,C^{0.5049}$ & $186.85\,C^{0.4319}$ \\
    \bottomrule
\end{tabular}